\documentclass[10pt,twocolumn,letterpaper]{article}

\usepackage[pagenumbers]{cvpr} %

\usepackage{multirow}
\usepackage{multicol}
\usepackage{makecell}
\usepackage{adjustbox}
\usepackage{bm}
\usepackage{xcolor} 
\usepackage{colortbl}
\usepackage[ruled,linesnumbered]{algorithm2e}
\usepackage[justification=justified,skip=3pt]{caption}
\usepackage{balance}
\usepackage{rotating}
\usepackage{soul}
\setul{0.8pt}{0.8pt}
\definecolor{cvprblue}{rgb}{0.21,0.49,0.74}
\usepackage[pagebackref,breaklinks,colorlinks,citecolor=cvprblue]{hyperref}

\makeatletter
\g@addto@macro\normalsize{%
  \abovedisplayskip 4pt plus 2pt minus 3pt%
  \belowdisplayskip \abovedisplayskip
  \abovedisplayshortskip 4pt plus2pt  minus3pt%
  \belowdisplayshortskip 4pt plus2pt minus3pt%
}

\makeatother

\def\paperID{14168} %
\def\confName{CVPR}
\def\confYear{2024}

\title{TEA: Test-time Energy Adaptation}

\author{Yige Yuan$^{12}$, Bingbing Xu$^{1}$, Liang Hou$^{4}$, Fei Sun$^{1}$, Huawei Shen$^{12}$, Xueqi Cheng$^{12}$\\
$^{1}$CAS Key Laboratory of AI Security, ICT, CAS\\
$^{2}$University of Chinese Academy of Sciences
$^{4}$Kuaishou Technology\\
{\tt\small \{yuanyige20z,xubingbing,sunfei,shenhuawei,cxq\}@ict.ac.cn, lianghou96@gmail.com}
}

\begin{document}
\maketitle
\begin{abstract}
Test-time adaptation (TTA) aims to improve model generalizability when test data diverges from training distribution, offering the distinct advantage of not requiring access to training data and processes, especially valuable in the context of large pre-trained models.
However, current TTA methods fail to address the fundamental issue: covariate shift, i.e., the decreased generalizability can be attributed to the model's reliance on the marginal distribution of the training data, which may impair model calibration and introduce confirmation bias.
To address this, we propose a novel energy-based perspective, enhancing the model's perception of target data distributions without requiring access to training data or processes.
Building on this perspective, we introduce \textbf{T}est-time \textbf{E}nergy \textbf{A}daptation (\textbf{TEA}), which transforms the trained classifier into an energy-based model and aligns the model's distribution with the test data's, enhancing its ability to perceive test distributions and thus improving overall generalizability.
Extensive experiments across multiple tasks, benchmarks and architectures demonstrate TEA's superior generalization performance against state-of-the-art methods.
Further in-depth analyses reveal that TEA can equip the model with a comprehensive perception of test distribution, ultimately paving the way toward improved generalization and calibration\footnote{Code is available: \url{https://github.com/yuanyige/tea}}. 
\end{abstract}
\vspace{-5mm}

\section{Introduction}
\label{sec:intro}

\begin{figure}[t]
    \centering
    \includegraphics[width=0.9\linewidth]{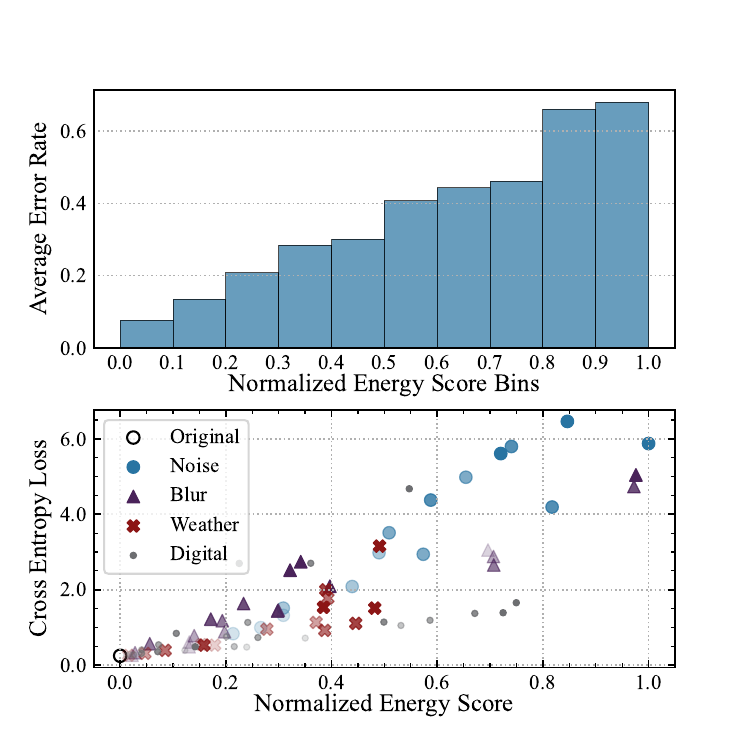}
    \caption{Performance vs. energy on model trained with original distribution, tested across various shifted distributions. \textbf{Upper}: error rate change within energy score groups. \textbf{Lower}: loss variation with energy scores, each point denoting a distribution. Marker styles and opacity reflect distribution types and divergence.}
    \label{fig:motivation}
    \vspace{-2mm}
\end{figure}

Deep neural networks, despite their remarkable performance under the assumption of independent and identically distributed (i.i.d.) training and test data~\cite{lecun2015deep,jordan2015machine,vapnik1999nature}, significantly degrade in real-world scenarios where unseen test data diverges from the training distribution. 
This limitation, known as distribution shift or domain shift, emphasizes the pressing need for generalizability across shifted test distributions~\cite{quinonero2008dataset, taori2020measuring,wang2022generalizing,yuan2022towards}.
To tackle this issue, recent studies of Test-Time Adaptation (TTA)~\cite{liang2023comprehensive,zhao2023on,wang2021tent} began considering the incorporation of unlabeled test data and leveraging it to fine-tune the source model in an unsupervised manner. 
This paradigm has garnered significant attention due to its ability to operate without access to training data or involvement with the training process.
In the era of large open-source models~\cite{wolf2019huggingface,touvron2023llama,du2021glm}, where models are publicly available but the training data and training process remain inaccessible due to privacy and resource restrictions~\cite{jain2016big,de2012data}, TTA emerges as especially beneficial and practical.

Existing TTA methods can be broadly categorized into three classes~\cite{liang2023comprehensive}.
Normalization-based methods~\cite{schneider2020improving,mirza2022norm} adjust the BatchNorm statistics of the model with test data statistics.
Entropy-based methods~\cite{wang2021tent,niu2022efficient,niu2023towards} fine-tune the model by minimizing the prediction entropy.
Pseudo-labeling-based methods~\cite{lee2013pseudo,liang2020we} utilize test-time generated labels for updates. 
While these methods have been empirically effective, these methods all fail to address a fundamental issue: covariate shift. Specifically, the decrease in generalization ability on test data  with distribution shift can be attributed to the model’s reliance on the marginal distribution of the training data. However, previous TTA methods do not address this shift due to their lack of connection with marginal distributions, impairing model calibration~\cite{guo2017calibration,jang2023testtime} and introducing confirmation bias~\cite{arazo2020pseudolabeling}.

To combat the above challenges, we propose a novel way rooted in an energy-based perspective. Within this way, energy is defined as an unnormalized probability assigned to a sample, where a lower score corresponds to a higher likelihood of that sample within a distribution~\cite{lecun2006tutorial,song2021train}. 
Proposing such a way to improve test time adaptation is twofold.

Firstly, test samples that correspond to \textbf{lower energy} within the model's distribution tend to exhibit \textbf{higher performance}. 
This is demonstrated by examining the energy scores of various test datasets in relation to a model trained on a specific training distribution. 
As depicted in \cref{fig:motivation}, an increase in the divergence between the test and training distributions is accompanied by a drastic escalation in energy scores, leading to a significant degradation in performance.
Secondly, the energy-based way can \textbf{address covariate shift} under TTA via directly injecting the model with a comprehensive \textbf{perception of test distribution}.
Addressing covariate shift in TTA is particularly challenging, as it is neither feasible to access the training dataset to align the marginal distribution between training and testing~\cite{shimodaira2000improving}, nor possible to modify the training process to mitigate the influence of marginal training distribution~\cite{NEURIPS2022_5ddcfaad}.
Under such circumstances, energy-based way can directly manipulate the trained model's likelihood landscape~\cite{lin2020likelihood} via an implicit distribution modeling process without requiring training process and training data, becoming a promising way.
This stands in contrast to other models such as GANs~\cite{goodfellow2020generative,hou2022augmentation}, Flows~\cite{rezende2015variational}, and VAEs~\cite{kingma2014autoencoding} which are advantageous only when the training data are accessible.

Building on the above energy-based way, we propose \textbf{T}est-time \textbf{E}nergy \textbf{A}daptation, abbreviated as \textbf{TEA}, which constructs an energy-based model from the trained (classifier) model by reinterpreting the negative log-sum-exp of logits as an energy function, and employs Contrastive Divergence~\cite{hinton2002training} as the adaptation objective to decrease the energy of test samples while increase that of samples generated by Stochastic Gradient Langevin Dynamics~\cite{welling2011bayesian}. 
This approach prevents a trivial solution that indiscriminately reduces the energy across the entire data space to ensure an increased likelihood for target test samples within the model's distribution.
TEA enables a gradual alignment between the distributions of the trained model and the test data, bolstering the trained model's perception of the test distribution and paving the way for superior adaptability and performance when confronted with the corresponding test data.

We investigate the effectiveness of TEA under image corruption and domain generalization on four popular benchmarks CIFAR-10, CIFAR-100, TinyImageNet and PACS, across three architectures WRN-28-10, ResNet-50 and ResNet-18. 
Experimental results underscore that TEA significantly outperforms current best-performing TTA methods in terms of generalizability, with an average increment of 4.7\%.
We further reveal that TEA can equip the model with a comprehensive perception of the test distribution. This, in turn, significantly improves the generalization and calibration of the trained model.

Our main contributions include:
\begin{itemize}
    \item \textbf{Promising Way}: We propose a new energy-based way for test time adaptation, which marks a departure from traditional methods and sheds light on potential avenues for mitigating the impact of distribution shifts.
    \item \textbf{Innovative Method}: We propose TEA to decrease the energy of the test data within the model's distribution, thereby equipping the trained model with a perception of the test distribution and enhancing generalizability. 
    \item \textbf{Extensive Experiments}: Experiments across extensive settings validate TEA's superiority over current leading methods. Further in-depth analyses extend the understanding of how energy reduction enhances the model's perception of test distribution, ultimately paving the way toward improved generalization and calibration.
\end{itemize}

\section{Related Work}

\begin{figure*}[t]
    \centering
    \includegraphics[width=0.96\linewidth]{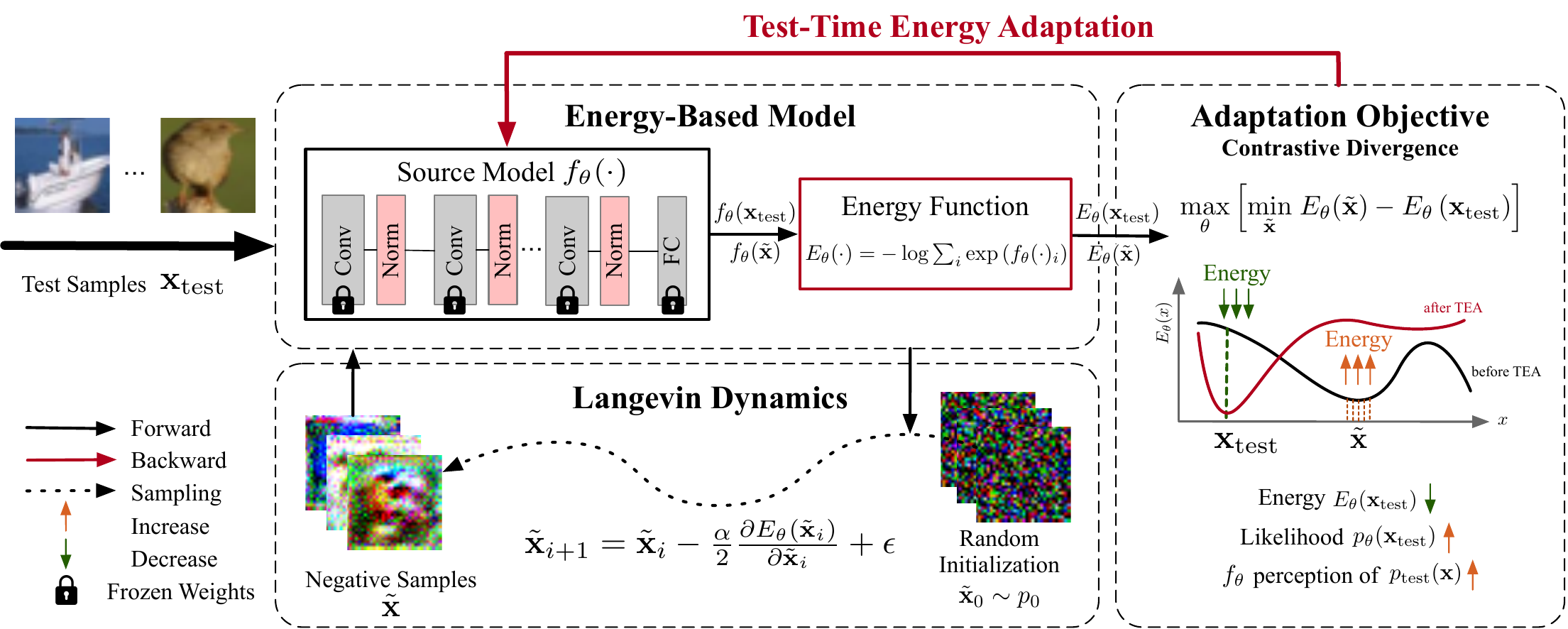}
\caption{\textbf{Overview of Test-time Energy Adaptation (TEA)}. 
Given a trained model (classifier) and in-coming test data, TEA directly integrates test data distribution into the trained classifier by fine-tuning its normalization layers through energy-based training: 
TEA constructs an \underline{Energy-Based Model} from the classifier by reinterpreting the negative log-sum-exp of logits as an energy function, and employs \underline{Contrastive Divergence} as the adaptation objective which decrease the energy of test samples while increase the energy of negative samples generated by \underline{Langevin Dynamics}. 
This adaptation increases the likelihood of test samples under the classifier's distribution, enabling a gradual alignment between the distributions of the trained classifier and the test data, thereby enhancing generalizability.}
\label{fig:model}
\end{figure*}

\paragraph{Test Time Adaptation}

Test Time Adaptation (TTA)~\cite{liang2023comprehensive} is a paradigm aiming to enhance a model's generalizability on specific test data through unsupervised fine-tuning with these data. 
Note that the model is originally trained on a distinct training dataset, which is not available during the adaptation phase.
Approaches like TTT~\cite{pmlr-v119-sun20b} adapt models through self-supervised proxy task during testing but require the training of the same proxy task during training procedure. 
DDA~\cite{Gao_2023_CVPR, xiao2023energybased} explores adapting the test data, yet faces limitations due to model structure and training constraints.
Recent research~\cite{wang2021tent} highlights a scenario where the training process and training data is entirely agnostic, leading to three main categories of approaches:
For normalization-based,
BN~\cite{schneider2020improving} adapts the BatchNorm~\cite{ioffe2015batch} statistics with test data.
DUA~\cite{mirza2022norm} uses a tiny fraction of test data and its augmentation for BatchNorm statistics adaptation.
For entropy-based, 
TENT~\cite{wang2021tent} fine-tunes BatchNorm layers using entropy minimization during the test phase.
EATA~\cite{niu2022efficient} employs a Fisher regularizer to limit excessive model parameter changes.
SAR~\cite{niu2023towards} removes high-gradient samples and promotes flat minimum weights.
For pseudo-labeling-based, 
PL~\cite{lee2013pseudo} fine-tunes parameters using confident pseudo labels.
SHOT~\cite{liang2020we} combines entropy minimization methods with pseudo labeling.

\paragraph{Energy Based Model}

Energy-Based Models (EBMs) are a type of non-normalized probabilistic models. Unlike most other probabilistic models, 
EBMs do not necessitate the normalizing constant to be tractable~\cite{lecun2006tutorial,song2021train} and do not require an explicit neural network for sample generation, implying the generation process is implicit~\cite{du2019implicit}.
These lead to increased flexibility in parameterization and allow for modeling a wider range of probability distributions.
Due to their flexibility, EBMs can construct hybrid models with both discriminative and generative capabilities, integrating the generative competencies into discriminative models without sacrificing their discriminative capabilities~\cite{Grathwohl2020Your,du2019implicit,han2019divergence}. Among these, JEM~\cite{Grathwohl2020Your} is particularly representative, reinterpreting classifiers as an EBM and achieving impressive results in both classification and generation.

\section{Method}
\label{sec:method}

In this section, we detail our method, Test-time Energy Adaptation (TEA).
Initially, we present a thorough description of the notation and overall architecture in~\cref{sec:overall_arch}, after which we proceed to explain energy adaption (\cref{sec:method_obj}) and modulation parameters (\cref{sec:method_para}), respectively.
Furthermore, we engage in a discussion about the difference between our method and entropy-based adaptation in~\cref{sec:method_disc}.

\subsection{Notation and Overall Architecture}
\label{sec:overall_arch}

The labeled training dataset is denoted as $\{(\mathbf{x}_{\text{train}}, y_{\text{train}})\} \subset \mathcal{X} \times \mathcal{Y}$ and the unlabeled test data are represented by $\mathbf{x}_{\text{test}} \in \mathcal{X}$, where $\mathcal{X}$ and $\mathcal{Y}$ are data and label spaces.
The respective marginal distributions of the training and test data are given by $\mathbf{x}_{\mathrm{train}} \sim p_{\mathrm{train}}(\mathbf{x})$ and $\mathbf{x}_{\mathrm{test}} \sim p_{\mathrm{test}}(\mathbf{x})$. 
A classifier model trained on the training dataset and parameterized by $\theta$, is denoted as $f_{\theta}: \mathcal{X} \rightarrow \mathcal{Y}$. 
The data distribution learned by this trained classifier is denoted by $p_{\theta}(\mathbf{x})$, which will be referred to as the model distribution henceforth.

The overall framework of TEA is depicted in~\cref{fig:model}. 
The motivation behind TEA is rooted in the issue of covariate shifts~\cite{jiang2008literature}, where the degradation of model generalization on test data $\mathbf{x}_{\mathrm{test}}$ is attributed to the model’s reliance on the training distribution $p_{\mathrm{train}}(\mathbf{x})$. 
To overcome this issue without accessing the training data or training process, TEA directly integrates the test data distribution into the trained model.
TEA constructs an energy-based model from the trained classifier by reinterpreting the negative log-sum-exp of logits as an energy function.
Through this lens, TEA employs contrastive divergence~\cite{hinton2002training} as the adaptation objective, which serves to decrease the energy (increase the likelihood) of the test samples under the model distribution $p_{\theta}(\mathbf{x})$.
This adaptation enables the gradual alignment of distribution between the trained model and test data, thereby bolstering the model's perception of the test distribution and enhancing generalizability.
Following previous TTA methods, TEA freezes the majority of the model's parameters, permitting only minor adjustments for efficient adaptation.

\subsection{Energy Adaptation for Test Distribution}
\label{sec:method_obj}

Enhancing the model's perception of test distribution from an energy-based perspective involves two key steps: constructing the energy-based model and optimizing it. 

\textbf{Constructing the energy-based model.}
To achieve this, we first introduce the basic idea of energy-based models (EBMs).
EBMs~\cite{lecun2006tutorial,song2021train,Grathwohl2020Your}  represent a class of probabilistic models that are characterized by an energy function. 
Consider a sample $\mathbf{x} \in \mathbb{R}^D$, the energy function $E: \mathbb{R}^D \rightarrow \mathbb{R}$ maps each sample into an energy that can be considered as an unnormalized probability, with lower scores indicating higher likelihoods~\cite{du2019implicit}.
Consequently, the probability density $p(\mathbf{x})$, as defined by EBM, can be expressed using the Boltzmann distribution~\cite{lifshitz1980statistical}, as shown in \cref{eq:ebm}, where the partition function $Z = \int \exp(-E(\mathbf{x})) \, \mathrm{d}\mathbf{x}$ serves to normalize the probability density.
\begin{equation}
    p(\mathbf{x})=\frac{\exp \left(-E(\mathbf{x})\right)}{Z}.
    \label{eq:ebm}
\end{equation}

Constructing an energy-based model from a trained classifier $f_{\theta}$ is founded on the fundamental analysis that an energy-based framework inherently underlies any discriminative model~\cite{Grathwohl2020Your}. In this framework, the energy of one sample for a corresponding class can be represented as its logit produced by the trained classifier, denoted by $E_{\theta}(\mathbf{x},y) = -f_\theta(\mathbf{x})[y]$.
Therefore, the joint probability distribution of $\mathbf{x}$ and $y$ can be defined as,
\begin{equation}
p_\theta(\mathbf{x}, y) = \exp(f_\theta(\mathbf{x})[y])/{Z_\theta},
\label{eq:p1}
\end{equation}
then the distribution of $\mathbf{x}$ can be obtained by marginalizing over $y$, as shown below,
\begin{equation}
p_\theta(\mathbf{x}) = \sum_y p_\theta(\mathbf{x}, y) = \sum_y \exp(f_\theta(\mathbf{x})[y])/{Z_\theta}.
\label{eq:p2}
\end{equation}
By substituting \cref{eq:p2} into \cref{eq:ebm}, we can obtain the form of the energy function as follows:
\begin{equation}
E_\theta(\mathbf{x})=-\log \sum_y \exp \left(f_\theta(\mathbf{x})[y]\right)
\label{eq:lse_energy}.
\end{equation}
Following the aforementioned steps, we repurpose and reinterpret the logits produced by the trained classifier to establish an energy-based model and define the energy function as the negative log-sum-exp of the logits.

After the logits reinterpretation, we can construct a energy-based model for the test data $\mathbf{x}_{\mathrm{test}}$ using the trained classifier, where $ Z_{\theta} = \int \sum_y \exp \left(f_\theta(\mathbf{x})[y]\right) \, \mathrm{d}\mathbf{x}$.
\begin{equation}\hspace{-2mm}p_\theta(\mathbf{x}_{\mathrm{test}}){=}\frac{\exp \left(\!-E_{\theta}(\mathbf{x}_{\mathrm{test}})\right)}{Z_{\theta}}{=}\frac{\sum_y\! \exp \left(f_\theta(\mathbf{x}_{\mathrm{test}})[y]\right)}{Z_{\theta}}.
\label{eq:px_test}
\end{equation}

\textbf{Optimizing the energy-based model.} Optimizing ~\cref{eq:px_test} is challenging. Specifically, the partition function $ Z_{\theta}$ necessitates integration across the whole input space of $\mathbf{x}$, typically making it computationally intractable. 
Thus, direct maximizing the log-likelihood of test data $\log p_{\theta}(\mathbf{x}_{\mathrm{test}})$ presents significant difficulties when training the parameter $\theta$ of our energy-based model. 
To overcome this difficulty, 
we propose to use contrastive divergence~\cite{hinton2002training,carreira2005contrastive} via estimating the gradient of the log-likelihood,
\begin{equation}
\frac{\partial \log p_\theta(\mathbf{x}_{\mathrm{test}})}{\partial \theta}=\mathbb{E}_{\tilde{\mathbf{x}} \sim p_\theta}\left[\frac{\partial E_\theta\left(\tilde{\mathbf{x}}\right)}{\partial \theta}\right]-\frac{\partial E_\theta(\mathbf{x}_{\mathrm{test}})}{\partial \theta}.
\label{eq:cd}
\end{equation}

In \cref{eq:cd}, the notation $\tilde{\mathbf{x}} \sim p_{\theta}$ denotes a random sample drawn from the distribution over $\mathbf{x}$, which is defined by the model's distribution. 
The sampling procedure can be performed through Stochastic Gradient Langevin Dynamics (SGLD)~\cite{welling2011bayesian}, which iteratively generates samples by using the gradient information~\cite{nijkamp2019learning,Grathwohl2020Your}. 
In this context, $p_0$ represents an initial noise distribution, $\alpha$ denotes the step-size, and $t=0,1,\dots,T{-}1$ is the time step. 
After $T$ steps of updating, a fictitious sample, generated by the energy-based model governed by the classifier, can be obtained.
\begin{equation}
    \tilde{\mathbf{x}}_{t+1}=\mathbf{x}_t-\frac{\alpha}{2} \frac{\partial E_\theta\left(\tilde{\mathbf{x}}_t\right)}{\partial \tilde{\mathbf{x}}_t}+\alpha\boldsymbol{\epsilon}, \quad \boldsymbol{\epsilon} \sim \mathcal{N}(\boldsymbol{0}, \mathbf{I}),
    \label{eq:sgld}
\end{equation}
this sampling process in \cref{eq:sgld} essentially optimizes the sample by moving in the direction of energy reduction. 
Consequently, the objective of \cref{eq:cd,eq:sgld} can be fundamentally understood as a min-max game (\cref{eq:minmax}), which minimizes the energy derived from the incoming test samples while concurrently amplifying the energy of fictitious samples obtained via SGLD from the classifier's distribution. 
Significantly, the latter plays a pivotal role in preventing a collapse towards a trivial solution where energy is indiscriminately minimized throughout the entire space, rather than focusing on the target test data.
\begin{equation}
    \underset{\theta}{\max} \left[ \underset{\tilde{\mathbf{x}}}{\min} \; E_{\theta}(\tilde{\mathbf{x}}) - E_{\theta}\left(\mathbf{x}_{\mathrm{test}}\right) \right].
    \label{eq:minmax}
\end{equation}

\begin{algorithm}[t]
\caption{Test Time Energy Adaptation}
  \label{tab:alg}
  \KwIn{Pre-trained Classifier $f_{\theta}$; Test Samples $\mathbf{x}_{\mathrm{test}}$; Langevin Sampling Step Size $\alpha$; Langevin Sampling Time Steps $T$; Noise Distribution $p_0$; Adaptation Rate $\beta$; Adaptation Steps $N$}
  \KwOut{Predictions for all $\mathbf{x}_{\mathrm{test}}$}
  
  $E_\theta(\cdot) \leftarrow -\log \sum_y \exp \left(f_\theta(\cdot)[y]\right)$
  
  \For{$i \leftarrow  0,1,\dots,N-1$}{

    $\tilde{\mathbf{x}}_0 \leftarrow  \mathrm{sample}(p_0)$
    
    \For{$t \leftarrow  0,1,\dots,T-1$}{

        $\boldsymbol{\epsilon} \leftarrow \mathrm{sample}(\mathcal{N}(\boldsymbol{0}, \mathbf{I}))$
        
        $\tilde{\mathbf{x}}_{t+1}\leftarrow \tilde{\mathbf{x}}_t-\frac{\alpha}{2} \frac{\partial E_\theta\left(\tilde{\mathbf{x}}_t\right)}{\partial \tilde{\mathbf{x}}_t}+\alpha\boldsymbol{\epsilon}$

    }

    $\tilde{\mathbf{x}} \leftarrow  \tilde{\mathbf{x}}_{T-1}$

    $ \theta \leftarrow \theta -\beta  \nabla_{\theta} \left[ E_{\theta}(\mathbf{x}_{\mathrm{test}}) - E_{\theta}(\tilde{\mathbf{x}}) \right]$
  }
 \Return{$f_{\theta}(\mathbf{x}_{\mathrm{test}})$}
 
\end{algorithm}

By adapting through this objective, the classifier's distribution continually converges towards the test distribution.
The convergence point of the min-max game is reached when the energy of samples drawn from the classifier's distribution matches the energy of the test samples.
At this stage, the classifier reaches a low-energy steady state with respect to the test data. 
The likelihood landscape of the classifier then exhibits a higher probability for samples that are consistent with the test data distribution, and conversely, a lower probability for samples that deviate from it, leading to a comprehensive perception of the classifier towards this test distribution and ultimately enhancing its generalizability.
The pseudocode for TEA can be found in \cref{tab:alg}.

\subsection{Modulation Parameters}
\label{sec:method_para}
As outlined in \cref{eq:minmax}, TEA requires updating the parameters of the trained model using the aforementioned energy adaptation to adjust to test data. 
In line with previous methods, we opt to update the parameters of the normalization layer due to the following two factors:
(1) \textbf{Practicality and Efficiency}: 
In the era of large-scale models, the practice of fine-tuning a selected group of parameters has gained prominence~\cite{hu2022lora}.  
For both practicality and efficiency, it's crucial to avoid updating all parameters as this process would be excessively time-consuming. 
Note that the parameters of the normalization layer account for a modest 1\% of total model parameters~\cite{wang2021tent}, making their update become more manageable.
(2) \textbf{Direct impact on data distribution}: 
The parameters within normalization layers possess the potential to capture intrinsic features of data, thereby exerting direct influence upon the corresponding data distribution. 
As evidenced by~\cite{huang2017arbitrary}, simple modulation to the mean and variance in a generator's normalization layers can modify image style, underlining the normalization parameters' profound impact on data distribution.
This aligns well with our goal of manipulating the energy of test data to make it compatible with the model distribution.

\subsection{Discussion}
\label{sec:method_disc}

As entropy-based adaptation has been the representative adaptation method, we further discuss the difference between TEA and entropy-based adaptation.
Intriguingly, TEA may have a connection with entropy-based adaptation, given that the negative entropy $\operatorname{NegEntropy} = \sum_{i} x_i\log x_i$ is the convex conjugate of the free energy $\operatorname{LogSumExp} = \log\sum_{i}\exp(x_i)$, as established in the literature~\cite{nielsen2016guaranteed,boyd2004convex}. 
Despite this connection, entropy-based methods, which apply softmax normalization in the label space $y$ and strive to minimize entropy, can result in diminished uncertainty within classification probabilities, leading to compromised model calibration. 
In contrast, TEA, utilizing the log-sum-exp function within the data space $\mathbf{x}$ can not only effectively avoid the pitfalls associated with entropy-based methods, but also improve calibration by introduces uncertainty to each class. 
This conjecture has been substantiated through experiments (refer to \cref{sec:exp_calibration}).

\section{Experiment}
In the following sections, we compare TEA with state-of-the-art methods across various tasks, benchmarks and architectures.
Then, we delve into deeper understanding of our method by exploring its desirable properties and identifying significant components that contribute to its improvements.
Due to the space limitations, more comprehensive experiments including full results on corruptions and other analyses, are provided in Appendix \cref{app:exp}.

\begin{table*}[!t]
\renewcommand{\arraystretch}{0.92}
\caption{Comparisons of TEA and baselines for image corruption on CIFAR-10(C), CIFAR-100(C), and Tiny-ImageNet(C) using WRN-28-10 with BatchNorm. Accuracy and mCE are evaluated at the most severe level and across all levels with asterisk (*) indicating the results are taken from the original paper~\cite{tang2023neuro}. The best adaptation results are highlighted in \textbf{boldface}.}
\centering
\begin{adjustbox}{width=\textwidth}
\setlength{\tabcolsep}{1mm}
\begin{tabular}{cl ccccc ccccc ccccc l}
\toprule
\multicolumn{2}{c}{\multirow{4}{*}{\parbox[t]{1.8cm}{\centering WRN-28-10 \\ BatchNorm}}} &  \multicolumn{5}{c}{CIFAR-10(C)} & \multicolumn{5}{c}{CIFAR-100(C)} & \multicolumn{5}{c}{Tiny-ImageNet(C)} \\
\cmidrule(lr){3-7} 
\cmidrule(lr){8-12}
\cmidrule(lr){13-17}
& 
& Clean &  \multicolumn{2}{c}{Corr Severity 5} & \multicolumn{2}{c}{Corr Severity 1-5} 
& Clean &  \multicolumn{2}{c}{Corr Severity 5} & \multicolumn{2}{c}{Corr Severity 1-5} 
& Clean &  \multicolumn{2}{c}{Corr Severity 5} & \multicolumn{2}{c}{Corr Severity 1-5} \\
\cmidrule(lr){3-3} \cmidrule(lr){4-5} \cmidrule(lr){6-7} 
\cmidrule(lr){8-8} \cmidrule(lr){9-10} \cmidrule(lr){11-12}
\cmidrule(lr){13-13} \cmidrule(lr){14-15} \cmidrule(lr){16-17}
& & Acc ($\uparrow$)  & Acc ($\uparrow$) & mCE ($\downarrow$) & Acc ($\uparrow$) & mCE ($\downarrow$) & Acc ($\uparrow$)  & Acc ($\uparrow$) & mCE ($\downarrow$)& Acc ($\uparrow$) & mCE ($\downarrow$) & Acc ($\uparrow$)  & Acc ($\uparrow$) & mCE ($\downarrow$)& Acc ($\uparrow$) & mCE ($\downarrow$)\\
\midrule
\multicolumn{2}{c}{\cellcolor{pink!20} Source }  
& 94.77 & 56.47 & 100.00 & 73.45 & 100.00 
& 81.79 & 35.39 & 100.00 & 52.12 & 100.00 
& 63.19 & 21.21 & 100.00 & 34.13 & 100.00 &\\
\cmidrule(lr){1-17}
\multirow{2}*{Norm}
& BN~\cite{schneider2020improving}
& 93.97 & 79.56 & 52.65 & 85.63 & 60.00 
& 80.83 & 60.06 & 63.54 & 68.11 & 69.42 
& 45.04 & 27.74 & 93.42 & 34.27 & 100.96 \\
& DUA*~\cite{mirza2022norm}
&-& 80.10 & 50.78 &-&-
&-&-&-&-&-
&-&-&-&-&- \\
\cmidrule(lr){1-17}
\multirow{2}*{Pseudo} 
& PL~\cite{lee2013pseudo}    
& 93.75 & 51.42 & 106.98 & 72.62 & 99.37 
& 80.52 & 53.40 & 72.12  & 64.53 & 75.29
& 47.84 & 28.26 & 91.22  & 39.83 & 91.67 \\
& SHOT~\cite{liang2020we}  
& 93.25 & 74.77 & 63.19 & 82.35 & 72.61 
& 80.52 & 56.53 & 68.01 & 66.00 & 73.28 
& 47.95 & 29.14 & 90.16 & \textbf{40.01} & \textbf{91.41} \\
\cmidrule(lr){1-17}
\multirow{4}*{Entropy} 
& TENT~\cite{wang2021tent}
& 93.66 & 81.41 & 48.13 & 86.75 & 56.17 
& 80.14 & 63.09 & 59.42 & 69.47 & 67.80 
& 39.54 & 26.31 & 95.52 & 32.03 & 104.49 \\
& ETA~\cite{niu2022efficient} 
& 93.96 & 79.58 & 52.64 & 85.63 & 59.99 
& 80.65 & 59.82 & 64.52 & 67.17 & 72.40 
& 43.20 & 27.28 & 94.12 & 33.46 & 102.25 \\
& EATA~\cite{niu2022efficient}
& 93.96 & 79.59 & 52.62 & 85.64 & 59.98 
& 80.68 & 60.24 & 63.75 & 67.48 & 71.66 
& 43.42 & 27.28 & 94.09 & 33.47 & 102.24 \\
& SAR~\cite{niu2023towards} 
& 93.97 & 79.77 & 51.94 & 85.83 & 58.97 
& 80.84 & 62.95 & 59.37 & 70.01 & 65.99 
& 41.58 & 28.21 & 92.82 & 34.60 & 100.47 \\
\cmidrule(lr){1-17}
\cellcolor{gray!20} Energy & \cellcolor{gray!20}TEA 
& \cellcolor{gray!20} \textbf{94.09} & \cellcolor{gray!20}\textbf{83.34} & \cellcolor{gray!20}\textbf{43.69} & \cellcolor{gray!20}\textbf{87.88} & \cellcolor{gray!20}\textbf{52.00} 
& \cellcolor{gray!20} \textbf{80.88} & \cellcolor{gray!20}\textbf{65.10} & \cellcolor{gray!20}\textbf{56.18} & \cellcolor{gray!20}\textbf{71.22} & \cellcolor{gray!20}\textbf{63.74}
& \cellcolor{gray!20} \textbf{51.65} & \cellcolor{gray!20}\textbf{31.67} & \cellcolor{gray!20} \textbf{87.99}  &\cellcolor{gray!20} 39.96 & \cellcolor{gray!20} 92.12 \\
\bottomrule

\end{tabular}

\end{adjustbox}
\label{tab:corr_bn}
\end{table*}
\begin{table}[!t]
\renewcommand{\arraystretch}{1}
\caption{Comparisons for image corruption on CIFAR-10(C), CIFAR-100(C), and Tiny-ImageNet(C) using ResNet-50 with GroupNorm across all severity levels. Best results are in \textbf{boldface}.}
\centering
\begin{adjustbox}{width=\linewidth}
\setlength{\tabcolsep}{1.5mm}
\begin{tabular}{cl cc cc cc}
\toprule
\multicolumn{2}{c}{\multirow{2}{*}{\parbox[t]{1.8cm}{\centering ResNet50 \\ GroupNorm}}} &  \multicolumn{2}{c}{CIFAR-10(C)} & \multicolumn{2}{c}{CIFAR-100(C)} & \multicolumn{2}{c}{Tiny-ImageNet(C)} \\
\cmidrule(lr){3-4} 
\cmidrule(lr){5-6} 
\cmidrule(lr){7-8} 
& & Acc ($\uparrow$) & mCE ($\downarrow$) 
& Acc ($\uparrow$) & mCE ($\downarrow$) 
& Acc ($\uparrow$) & mCE ($\downarrow$) \\
\midrule
\multicolumn{2}{c}{\cellcolor{pink!20}Source} 
& 78.71 & 100.00 & 54.98 & 100.00 & 26.64 & 100.00 \\
\cmidrule(lr){1-8}
\multirow{2}*{Pseudo} 
& PL   & 79.43 & 94.76 & 56.68 & 96.02 & 26.60 & 99.92 \\
& SHOT & 81.98 & 86.65 & 58.31 & 93.45 & 29.11 & 96.73 \\
\cmidrule(lr){1-8}
\multirow{4}*{Entropy} 
& TENT & 77.29 & 102.88 & 56.34 & 96.88 & 26.65 & 99.94 \\
& ETA  & 78.68 & 100.09 & 56.72 & 96.37 & 29.25 & 96.42 \\
& EATA & 78.70 & 100.02 & 56.76 & 96.28 & 29.25 & 96.42 \\
& SAR  & 78.78 & 99.65  & 55.28 & 99.33 & 27.05 & 99.41 \\
\cmidrule(lr){1-8}
\cellcolor{gray!20} Energy 
& \cellcolor{gray!20}TEA  
& \cellcolor{gray!20} \textbf{83.05} 
& \cellcolor{gray!20} \textbf{79.09}
& \cellcolor{gray!20} \textbf{59.67}
& \cellcolor{gray!20} \textbf{89.32}
& \cellcolor{gray!20} \textbf{30.41}
& \cellcolor{gray!20} \textbf{94.81}
\\
\bottomrule

\end{tabular}

\end{adjustbox}
\label{tab:corr_gn}
\end{table}
\begin{table}[h]
\renewcommand{\arraystretch}{0.82}
\caption{Single source domain generalization comparisons on PACS datasets using ResNet-18 with BatchNorm in terms of Accuracy. The best adaptation results are highlighted in \textbf{boldface}.}
\centering
\begin{adjustbox}{width=\linewidth}
\setlength{\tabcolsep}{2.5mm}
\begin{tabular}{ccccccc}
\toprule
\multirow{2}{*}{\parbox[t]{1.8cm}{\centering Source\\ Domain}} & \multirow{2.5}{*}{Method} & \multicolumn{4}{c}{Target Domain} & \multirow{2.5}{*}{Avg} \\
\cmidrule(lr){3-6}
& & Photo & Art & Cartoon & Sketch & \\
\midrule
\multirow{8}{*}{Photo} 
& \cellcolor{pink!20} Source &-& 26.76 & 22.40 & 16.62 & 21.93 \\
\cmidrule(lr){2-7}
& BN     &-& 26.66 & 27.94 & 15.96 & 23.52 \\
& TENT   &-& 26.95 & 29.86 & 17.54 & 24.78 \\
& EATA   &-& 26.66 & 28.11 & 15.98 & 23.59 \\
& SAR    &-& 26.71 & 28.41 & 15.98 & 23.70\\
& SHOT   &-& 26.61 & 29.86 & 20.92 & 25.80 \\
\cmidrule(lr){2-7}
& \cellcolor{gray!20} TEA & \cellcolor{gray!20}- & \cellcolor{gray!20}\textbf{28.81} &\cellcolor{gray!20}\textbf{33.62} & \cellcolor{gray!20}\textbf{20.49} &\cellcolor{gray!20}\textbf{27.64}\\ %
\midrule
\multirow{8}{*}{Art} 
& \cellcolor{pink!20} Source & 49.04 &-& 36.43 & 24.48 & 36.65\\
\cmidrule(lr){2-7}
& BN     & 46.65 &-& 28.28 & 22.73 & 32.55 \\
& TENT   & 50.78 &-& 30.12 & 24.61 & 35.17 \\
& EATA   & 46.83 &-& 29.31 & 23.42 & 33.19\\
& SAR    & 47.90 &-& 33.02 & 26.27 & 35.73 \\
& SHOT   & 50.24 &-& 34.30 & \textbf{29.37} & 37.97\\
\cmidrule(lr){2-7}
& \cellcolor{gray!20}TEA & \cellcolor{gray!20}\textbf{56.29} & \cellcolor{gray!20}- & \cellcolor{gray!20}\textbf{38.57} & \cellcolor{gray!20}28.71 & \cellcolor{gray!20}\textbf{41.19}\\
\midrule
\multirow{8}{*}{Cartoon} 
& \cellcolor{pink!20} Source & 42.69 & 29.79 &-& 29.47 & 33.98 \\
\cmidrule(lr){2-7}
& BN     & 28.68 & 25.15 &-& 20.87 & 24.90 \\
& TENT   & 30.96 & 23.34 &-& 22.65 & 25.65 \\
& EATA   & 28.80 & 25.10 &-& \textbf{25.04} & 26.31 \\
& SAR    & 29.70 & 25.78 &-& 21.51 & 25.66 \\
& SHOT   & \textbf{37.72} & 22.66 &-& 23.14 & 27.84 \\
\cmidrule(lr){2-7}
& \cellcolor{gray!20}TEA & \cellcolor{gray!20}36.05 & \cellcolor{gray!20}\textbf{31.44} & \cellcolor{gray!20}- & \cellcolor{gray!20}22.88 & \cellcolor{gray!20}\textbf{30.12}\\
\midrule
\multirow{8}{*}{Sketch} 
& \cellcolor{pink!20} Source & 19.94 & 18.70 & 32.21 &-& 23.62 \\
\cmidrule(lr){2-7}
& BN     & 13.47 & 17.14 & 29.86 &-& 20.16 \\
& TENT   & 13.53 & 17.38 & 29.52 &-& 20.14 \\
& EATA   & 13.17 & 17.33 & 30.08 &-& 20.19 \\
& SAR    & 13.29 & 18.80 & 29.95 &-& 20.68 \\
& SHOT   & \textbf{19.76} & 18.75 & 30.46 &-& 22.99 \\
\cmidrule(lr){2-7}
& \cellcolor{gray!20}TEA & \cellcolor{gray!20}19.64 & \cellcolor{gray!20}\textbf{21.24} & \cellcolor{gray!20}\textbf{33.19} & \cellcolor{gray!20}- & \cellcolor{gray!20}\textbf{24.69} \\
\bottomrule
\end{tabular}
\end{adjustbox}
\label{tab:dg}
\end{table}
\vspace{-2mm}

\subsection{Experimental Setup}

\paragraph{Datasets and Metrics} 
We focus on two tasks to verify the performance of TEA: generalization on image corruption and domain generalization. 
Image corruption includes clean and corrupted datasets from CIFAR-10(C), CIFAR-100(C) and  TinyImageNet-200(C)~\cite{krizhevsky2009learning,le2015tiny,hendrycks2018benchmarking}, incorporating 15 types of corruption at five severity levels. 
Domain generalization considers the PACS dataset~\cite{li2017deeper}, encompassing four domains (Photo, Art Painting, Cartoon, Sketch) across seven categories.
We use Accuracy and mean Corruption Error (mCE)~\cite{hendrycks2018benchmarking} as evaluation metrics, adhering to the protocol in~\cite{yuan2023pde}. 
The evaluations are conducted at both the most severe level and the average of all severity levels to ensure thorough analysis. 
\paragraph{Backbones and Baselines} 
We use two architectures for image corruption: WideResNet-28-10~\cite{zagoruyko2016wide} with BatchNorm~\cite{ioffe2015batch}, and ResNet-50~\cite{he2016deep} with GroupNorm~\cite{wu2018group}, consistent with the implementations of TENT~\cite{wang2021tent} and SAR~\cite{niu2023towards}. 
We use ResNet-18~\cite{he2016deep} for domain generalization. 
We evaluate our method against eight leading TTA methods across  three categories: 
(1) Normalization-based methods: BN~\cite{schneider2020improving} and DUA~\cite{mirza2022norm}. 
(2) Entropy-based methods: TENT~\cite{wang2021tent}, ETA, EATA~\cite{niu2022efficient}, and SAR~\cite{niu2023towards}; 
(3) Pseudo-labeling-based methods: PL~\cite{lee2013pseudo} and SHOT~\cite{liang2020we}.
Source denotes the original model without any adaptation.
\paragraph{Implementation} 
We implement methods based on PyTorch~\cite{NEURIPS2019_9015}. Consistency in model weight is ensured by the RobustBench protocol \cite{croce2021robustbench}, which provides pre-trained weights for the WideResNet-28-10 (BatchNorm) on CIFAR-10. In the case where RobustBench weights are unavailable, we train models in accordance with the guidelines specified in \cite{zagoruyko2016wide}.
All adaptation employ Adam~\cite{DBLP:journals/corr/KingmaB14}, except for SAR, which originally uses SAM \cite{foret2021sharpnessaware} with SGD~\cite{robbins1951stochastic}. 
Baselines are replicated using their original hyper-parameters, except when these were unspecified. More details and setups are deferred to Appendix~\cref{app:setting}.

\subsection{Adaptation Results}
\label{sec:exp_main}

In this section, we evaluate the generalizability of TEA in comparison to state-of-the-art methods across two tasks including image corruption and domain generalization.

\paragraph{Image Corruption}
As reported in \cref{tab:corr_bn}, we conducted experiments on three benchmarks against eight baselines for corruption scenarios, with ``*'' indicating results that are taken from the original paper~\cite{tang2023neuro}.
TEA markedly surpasses all baselines in the vast majority of datasets and severity levels. Specifically, TEA outshines the best-performing baseline by an average of 4.7\% at the most severe level.
The only exception is on Tiny-ImageNet across all levels, where TEA ranks second, merely trailing by a minimal margin  0.1\%.
To provide a broader validation of TEA's performance, we further incorporate the ResNet50 with GroupNorm. As indicated in~\cref{tab:corr_gn}, TEA maintains its leading position, delivering the best performance in both average accuracy and mCE.
These results underscore the efficacy of TEA in handling image corruption scenarios, ensuring its universality for model architecture or normalization techniques.

\begin{figure*}[!t]
    \centering
    \includegraphics[width=1\linewidth]{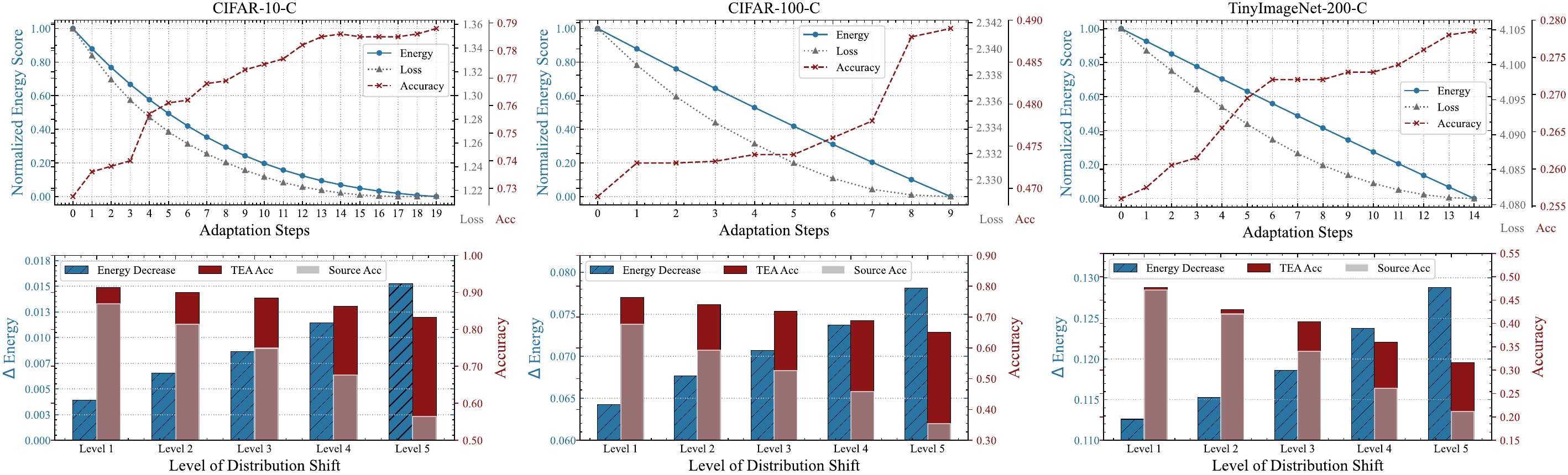}
    \caption{This illustration captures the energy reduction and generalizability enhancement achieved by TEA across CIFAR-10-C, CIFAR-100-C, and TinyImageNet-200-C, displayed from left to right. The \textbf{upper} set of graphs trace the evolution of energy score, corresponding loss and accuracy in response to incrementally increasing TEA adaptation steps. The \textbf{lower} set uncovers the extent of energy reduction and the consequent performance improvement before and after executing TEA adaptation, under different levels of distribution shift.}
    \label{fig:energy_reduce}
\end{figure*}

\vspace{-3mm}
\paragraph{Domain Generalization}
\cref{tab:dg} provides a comparison between TEA and state-of-the-art TTA approaches on the PACS dataset. 
It is evident that when trained on the photo and art domains, TEA exhibits substantial improvements over the best-performing baseline, achieving increases of 7.1\% and 8.5\%, respectively. 
Compared to photo and art, cartoon and sketch domains may exhibit greater domain discrepancies, posing significant challenges to model generalization.
Despite these challenging conditions, in the adaptation from cartoon to art and the sketch domain, TEA achieved improvements when other baselines all experienced significant declines. E.g., TENT shows decline of 14.73\% compared to the source in sketch, whereas TEA bucks the trend and enhances performance by 4.5\%, highlighting its stability in the face of severe domain shifts.

The results from both tasks highlight TEA's superior generalizability.
These effectiveness may originate from the reduced energy, the enhanced distribution perception, and the improved calibration. 
In next section, we will delve into these aspects for further analysis and discussion.

\subsection{Analysis and Discussion}

In this section, we delve into the mechanisms driving TEA's effectiveness and explore its desirable properties.
Specifically, we studied three key aspects of TEA:
(1) The correlation between energy reduction and generalizability enhancement;
(2) The distribution perception and generation capabilities;
(3) The confidence calibration improvements.

\vspace{-2mm}
\subsubsection{Relation between TEA's Energy Reduction and Generalizability Enhancement} 
\label{sec:exp_energy}

This experiment validate TEA's energy reduction capability, and its impact on generalizability, spanning two scenarios:
(1) In the first scenario, we focus on the trends of energy scores, loss, and accuracy on the same test data over increasing adaptation steps.
(2) In the second scenario, we explore how the extent of energy reduction correlates with performance improvements, before and after adaptation, using varied test data that exhibit different distribution shifts.

The results from both scenarios are depicted in \cref{fig:energy_reduce}. 
It is observable that 
$(\romannumeral1)$ As the iteration step of TEA increases, there is a consistent reduction in energy and corresponding decrease in loss, coupled with an ongoing enhancement in accuracy. 
$(\romannumeral2)$ As the distribution shift increases, TEA's energy reduction becomes more pronounced.  Concurrently, the enhancement in performance over the baseline also increases.
$(\romannumeral3)$ As the distribution shift increases, there is a sharp degradation observed in the baseline performance. However, the model adapted via TEA maintains its stability and robustness, demonstrating resilience against strong distribution shifts.
In summary, these aforementioned trends are consistently observed across three datasets, demonstrating TEA's significant effectiveness in reducing energy. 
Notably, a greater reduction in energy correlates with an increased improvement in the model's generalizability

\subsubsection{TEA's Distribution Perception and Generation} 
\begin{figure}[t]
    \centering
    \includegraphics[width=1\linewidth]{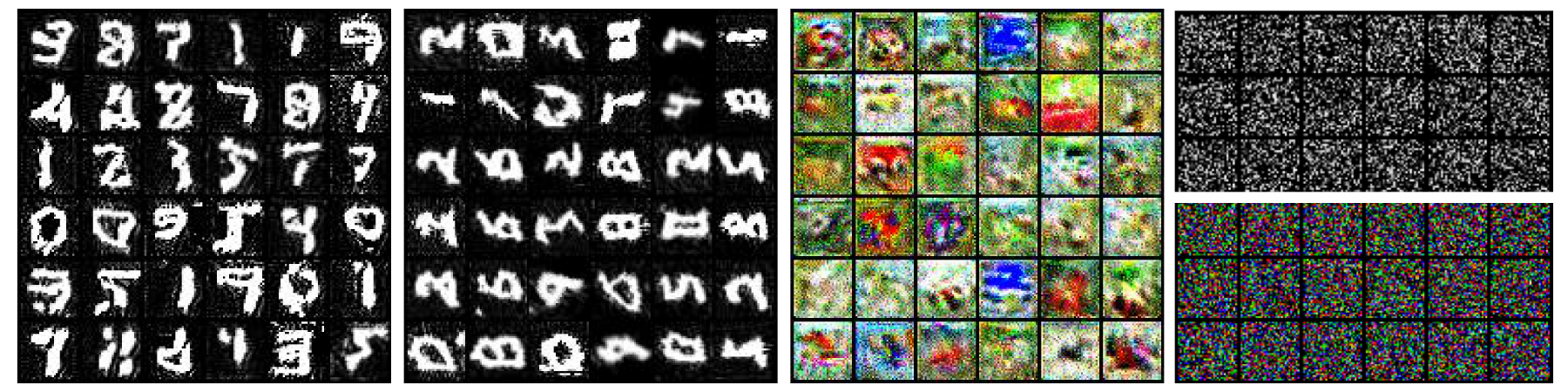}
    \caption{Test distribution perception visualization for identical training and testing distributions on MNIST and CIFAR-10.}
    \label{fig:vis_mnist}
    \centering
    \includegraphics[width=1\linewidth]{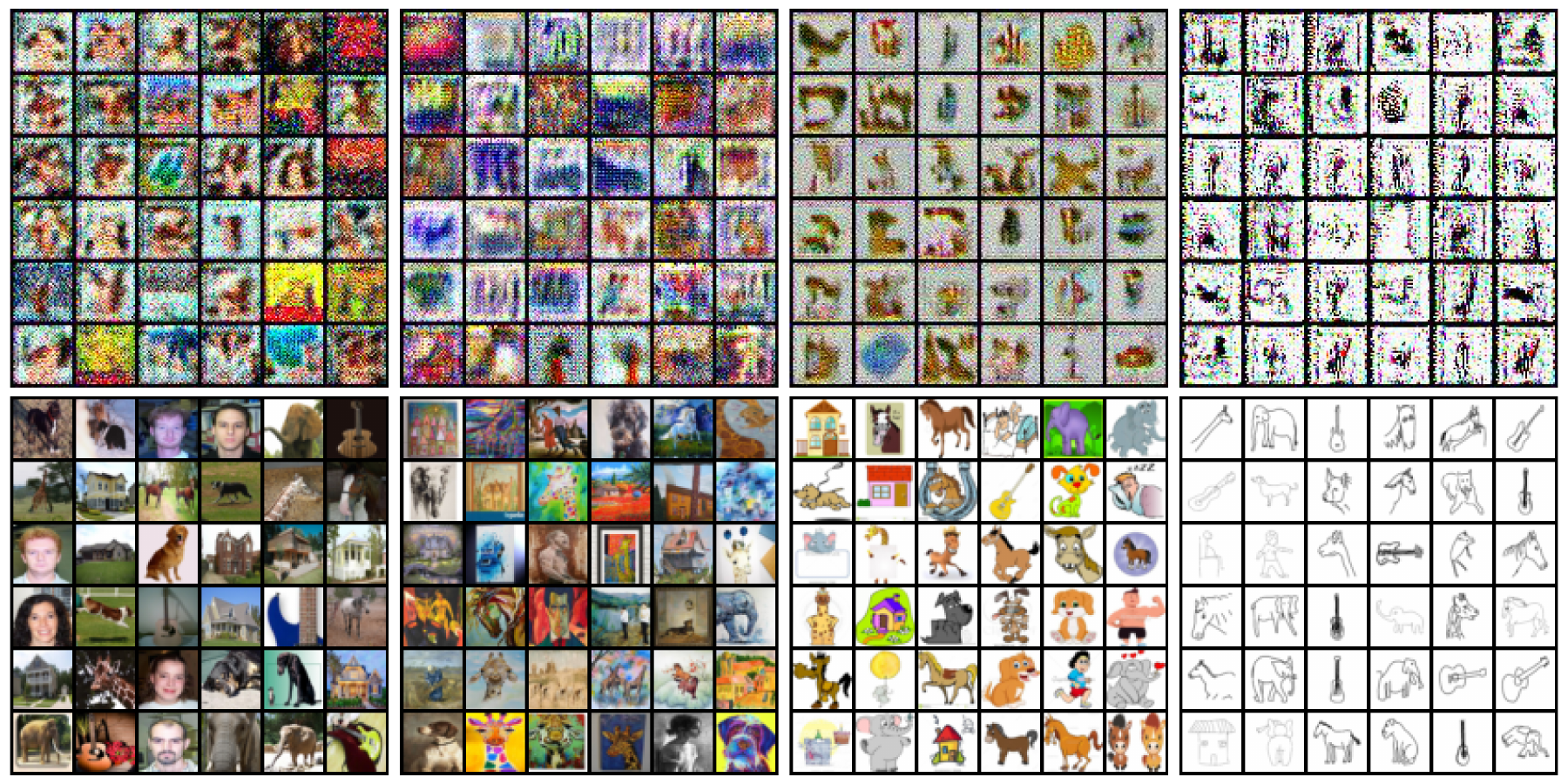}
    \caption {Test distribution perception visualization (\textbf{upper}) and real samples (\textbf{lower}) on shifted distribution: A model trained on PACS-A dataset then individually tested with TEA adaptation across PACS-P, PACS-A, PACS-C, PACS-S datasets.}
    \label{fig:vis_pacs}
\end{figure}

This experiment aims to validate the capability of TEA in perceiving and modeling test data distribution. 
We framed the experiment within two scenarios: identical training and testing distributions, and shifted training and testing distributions.
(1) The first scenario include four settings: 
$(\romannumeral1)$ Source model on MNIST training set, TEA adaptation on MNIST test set;
$(\romannumeral2)$ Source model on MNIST training set, TEA adaptation on 90-degree rotated MNIST test set;
$(\romannumeral3)$ Source model on CIFAR-10 training set, TEA adaptation on CIFAR-10 test set;
$(\romannumeral4)$ Source models on MNIST or CIFAR-10 training sets, tested without TEA adaptation.
(2) In the second scenario, source model was trained on the PACS-A dataset and individually tested with TEA adaptation on PACS-P, PACS-A, PACS-C and PACS-S datasets.

The outcomes from both scenarios are respectively illustrated in \cref{fig:vis_mnist} and \cref{fig:vis_pacs}. \cref{fig:vis_mnist} (1,3) indicates that, in scenarios where the distributions are closely identical, TEA has the potential to reconstruct samples that maintain discernible patterns.
\cref{fig:vis_mnist} (2) reveals that the sample distribution indeed originates from the test data rather than provoking recollections from the training datasets.
\cref{fig:vis_mnist} (4) confirms that models without adaptation, or those adapted using other methods, fail to effectively characterize the testing distributions on both the simple MNIST dataset (upper) and the more complex CIFAR-10 dataset (lower).
Drawing conclusions from \cref{fig:vis_pacs}, we can infer that under significant distribution/domain shift, our method can still characterize the key features of the shifted test distribution, such as style, texture and color schemes.

In summary, our approach endows the model with generative ability for test data via energy-based fine-tuning of the normalization layers. 
This ability may explain the source of TEA's generalizability: it incorporates generative self-supervised information from the test data into the model and improves the model's thorough understanding of the test distribution, which in turn strengthens its generalization performance on that distribution.

\vspace{-2mm}
\subsubsection{TEA's Improvements in Confidence Calibration} 
\label{sec:exp_calibration}

This experiment compares model calibration across the source model, the entropy-based method TENT, the pseudo-labeling-based method SHOT, and our energy-based TEA, on CIFAR-10.
In accordance with the protocol in~\cite{guo2017calibration,chen2022contrastive}, we illustrate reliability histogram and compute two scalar summary statistics: Expected Calibration Error (ECE) and Maximum Calibration Error (MCE)~\cite{guo2017calibration}, to evaluate calibration implemented by torchmetrics. 
The procedures are implemented as follows: For the reliability histogram, we divided the model's predictions into ten bins based on the confidence score of the highest probability class and calculated the average accuracy for each bin.

\begin{figure}[!t]
\centering
\includegraphics[width=\linewidth,height=0.33\textheight]{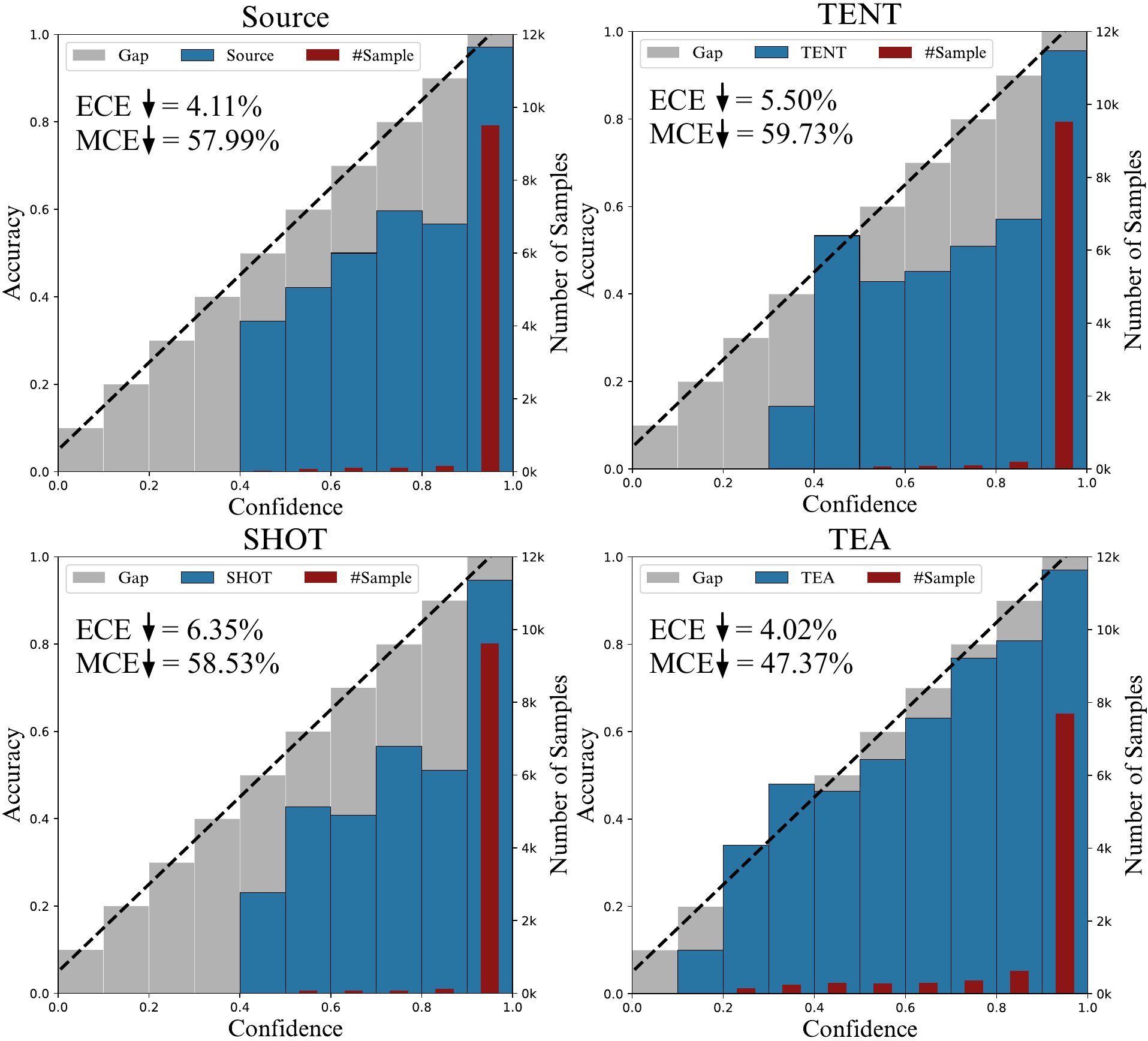} %
\caption{Calibration comparison between TEA and baselines on CIFAR-10 dataset. In an ideal scenario for optimal calibration, blue bars should align with the diagonal line, and a smaller grey gap area is preferred. Quantitative measures are provided via ECE and MCE metrics, where lower values indicate better calibration.}
\label{fig:calibration}
\end{figure}

The results are depicted in \cref{fig:calibration}. 
From the perspective of the histogram, an optimally calibrated model should have its bar graph in a diagonal shape to achieve the smallest gap area.
However, the bars for TENT and SHOT are observed to fall significantly below this line, manifesting an even inferior performance compared to the source model without any adaptation. These phenomena provide evidence that both the entropy-based methods and the pseudo-label-based method could potentially harm the confidence calibration by inducing overconfidence in their predictions.
In stark contrast, our TEA has significantly narrowed the gap area and improved alignment with the diagonal line. For quantitative metrics, ECE and MCE, TEA has improved by 2.43\% and 18.31\% respectively compared to the source model.

The improvement in calibration of TEA over competitors may come from the following reason: Neural networks inherently tend to be overconfident~\cite{guo2017calibration}. The softmax function enforces exponential normalization among classes, which tends to amplify the probability of dominant classes, thus inherently not advantageous for calibration.
Methods like TENT and SHOT exacerbate this dominance of certain classes by reducing the uncertainty of class probabilities, further amplifying the overconfidence of the classifier.
On the contrary, TEA dose not perform normalization in the label space, but maximizes the log-sum-exp of classifier logits, which essentially introduces a certain level of uncertainty to each class and empowers TEA with the ability to enhance calibration.

\section{Conclusion}
To achieve test-time adaptation, we introduce an innovative energy-based perspective to mitigate the impact derived from distribution shifts. %
The proposed TEA aims to decrease the energy of the test data within the pre-trained model's distribution. 
TEA guides the model towards achieving a harmonious low-energy equilibrium state for the test data, which mitigates the model's distribution discrepancy and boosts its generalizability towards test distributions.
Comprehensive experiments across multiple tasks, benchmarks, and architectures confirm TEA’s superiority over current leading methods. Further in-depth analyses of TEA’s underlying mechanisms deepen our understanding of how energy reduction can enhance the model's perception of the test distribution, ultimately paving the way for improved generalization and calibration.

{
    \small
    \bibliographystyle{ieeenat_fullname}
    \bibliography{main}
}

\clearpage
\setcounter{page}{1}
\maketitlesupplementary

\section{Appendix Summary}
The appendix contains the following sections:
\begin{itemize}
    \item [(1)] Additional Experiments and Analyses (\cref{app:exp}):
    \begin{itemize}
        \item Detailed Results for Energy Reduction (\cref{app:full_energy})
        \item Detailed Results for Image Corruption (\cref{app:full_corr})
        \item Hyper-parameters Sensitivity (\cref{app:exp_hyper})
    \end{itemize}
    \item [(2)] Detailed Settings (\cref{app:setting}):
    \begin{itemize}
        \item Datasets(\cref{app:datasets})
        \item Evaluation Metrics(\cref{app:metrics})
        \item Hyper-parameters(\cref{app:hyper})
        \item Computing Resources (\cref{app:resources})
    \end{itemize}
    \item [(3)] Limitations and Future Explorations (\cref{app:limitation}).
\end{itemize}

\section{Additional Experiments}
\label{app:exp}

\subsection{Detailed Results for Energy Reduction}
\label{app:full_energy}
\begin{figure*}[t]
    \centering
    \includegraphics[width=1\linewidth]{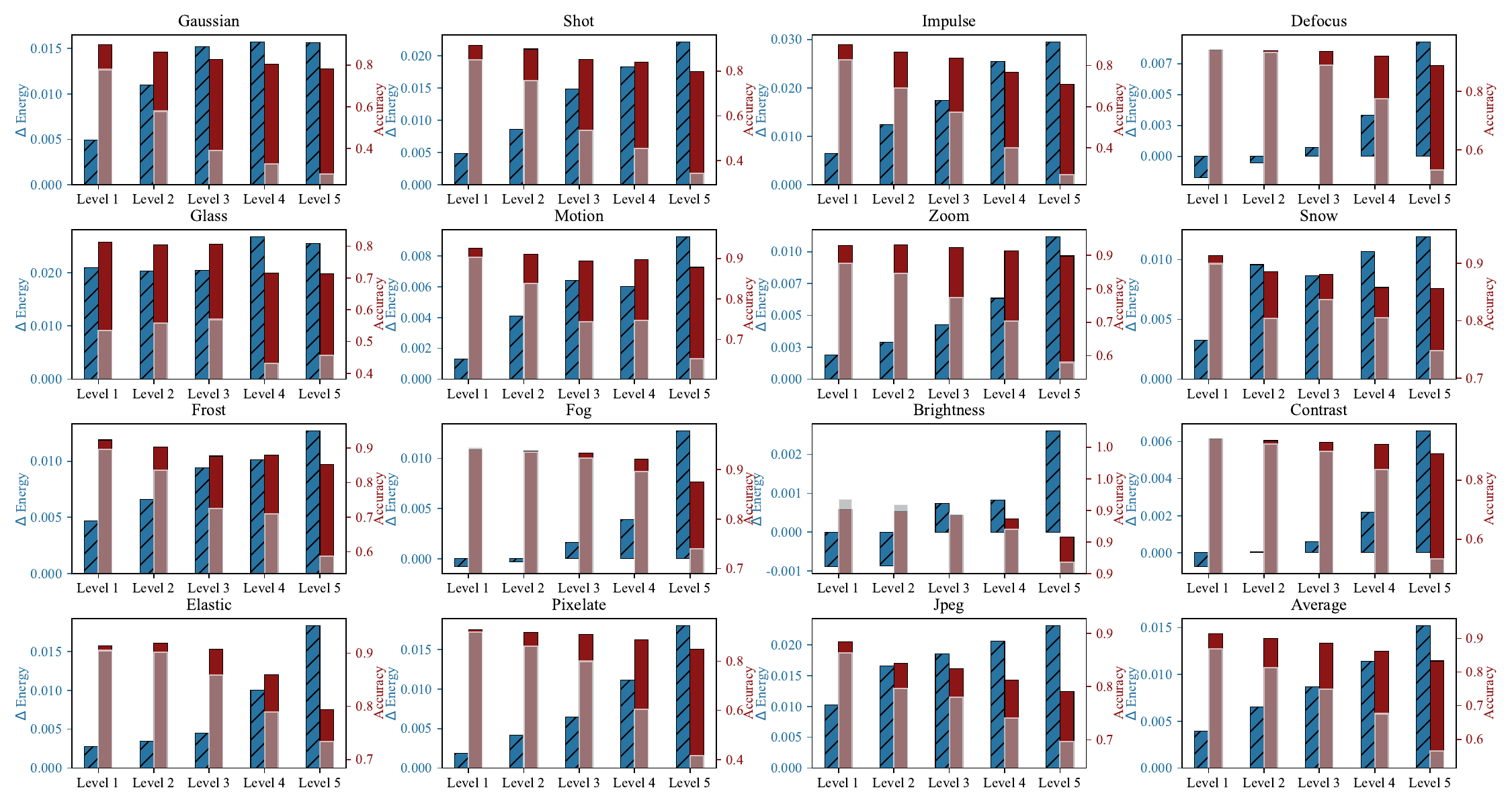}
    \caption{The relationship between TEA's energy reduction and the enhancement of generalizability on CIFAR-10-C, under different types of distribution and different severity level of distribution shifts. Each subfigure plots corruption severity level on the x-axis, energy reduction on the left y-axis, and accuracy on the right y-axis. The accuracy axis contains two bars: the red bar denotes our TEA' accuracy, while the transparent bar denotes baseline's accuracy.}
    \label{fig:app_energy_reduce_full1}
    \includegraphics[width=1\linewidth]{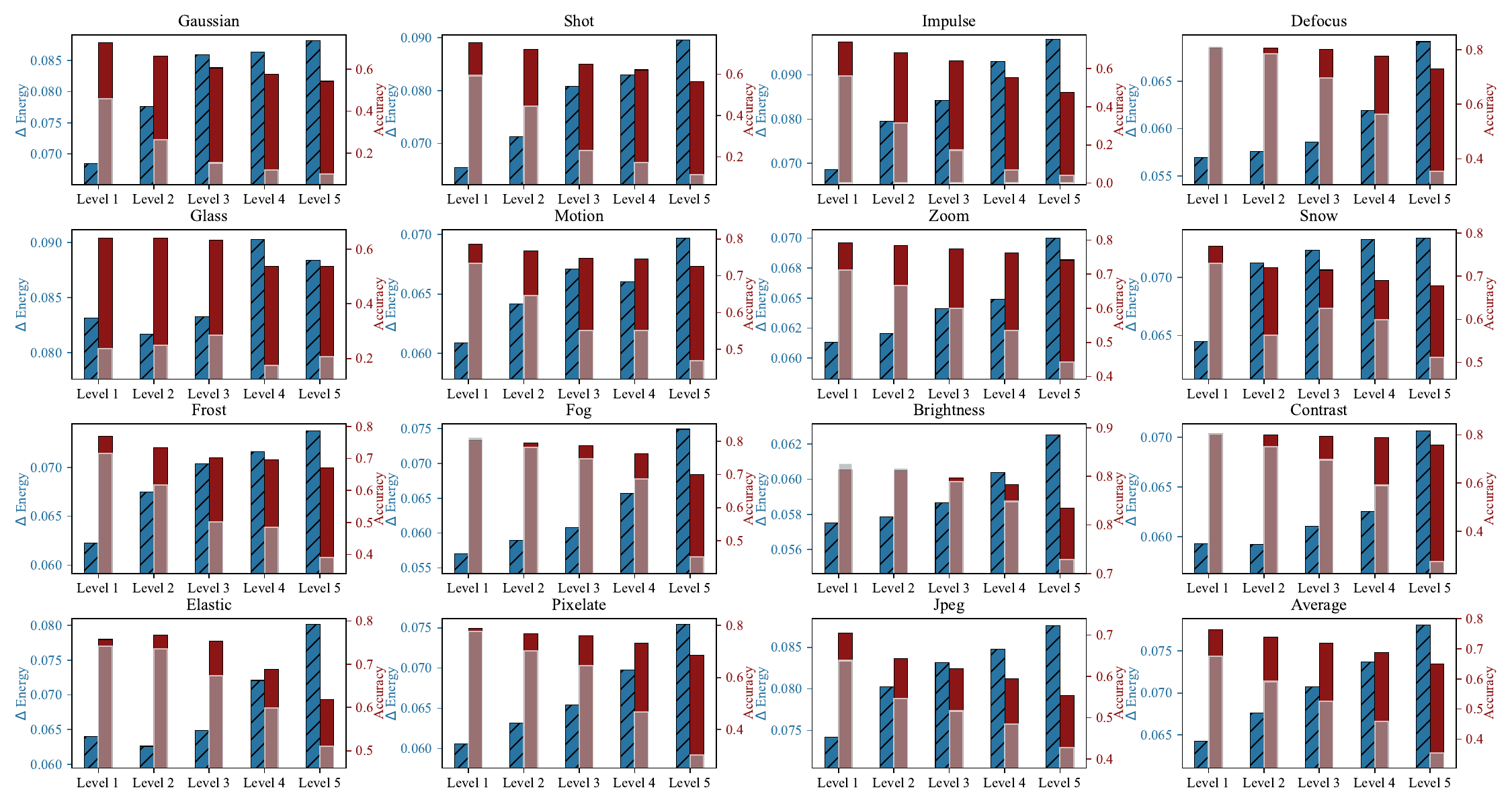}
    \caption{The relationship between TEA's energy reduction and the enhancement of generalizability on CIFAR-100-C, under different types of distribution and different severity level of distribution shifts. Each subfigure plots corruption severity level on the x-axis, energy reduction on the left y-axis, and accuracy on the right y-axis. The accuracy axis contains two bars: the red bar denotes our TEA' accuracy, while the transparent bar denotes baseline's accuracy.}
    \label{fig:app_energy_reduce_full2}
\end{figure*}

This section serves as an extension of the energy analysis (\cref{sec:exp_energy}) in the main text, presenting the relationship between TEA's energy reduction and the enhancement of generalizability across all types of corruption. 
The detailed results are shown in \cref{fig:app_energy_reduce_full1,fig:app_energy_reduce_full2}, where each corruption type is analyzed at five levels of severity, with the analysis examining the correlation between the extent of energy reduction and performance improvements, both before and after adaptation, as severity levels increase. 

In our experiments, TEA generally reduced energy and enhanced generalization across various corruptions. 
Yet, for mild corruptions like ``Brightness'' at level one, i.e., the mildest in CIFAR-10-C, generalization did not improve and occasionally deteriorated slightly. 
Correspondingly, energy did not decrease and even increased marginally. 
These outcomes indicate a strong correlation between generalizability enhancement and energy reduction. 
However, it is possible that our method may not reduce energy as anticipated for distributions with some less severe corruption types.
This may be attributed to these distributions being closely aligned with the original, already at a low energy state. 
The uniform hyperparameters used in our adaptation may not be optimal for such cases. 
Addressing this discrepancy will be a priority in future research.

\subsection{Detailed Results for Image Corruption}
\label{app:full_corr}
This section serves as an extension of the main adaption results (\cref{sec:exp_main}) in the main text, presenting the detailed performance for each corruption type at the most severe corruption level.
The detailed results are shown in \cref{tab:app_mainada_full}.
In our evaluation, TEA consistently achieves the highest accuracy for every corruption type on CIFAR-10-C and CIFAR-100-C datasets. On Tiny-ImageNet, our model exhibits superior performance on the majority of corruptions. However, it is slightly outperformed by SHOT on a few corruption types. 
The performance difference might be because the corruptions are mild and similar to the source data, which benefits pseudo-label methods like SHOT that rely on this similarity to produce accurate labels.

\subsection{Hyper-parameters Sensitivity}
\label{app:exp_hyper}
\begin{figure*}[t]
    \centering
    \includegraphics[width=1\linewidth]{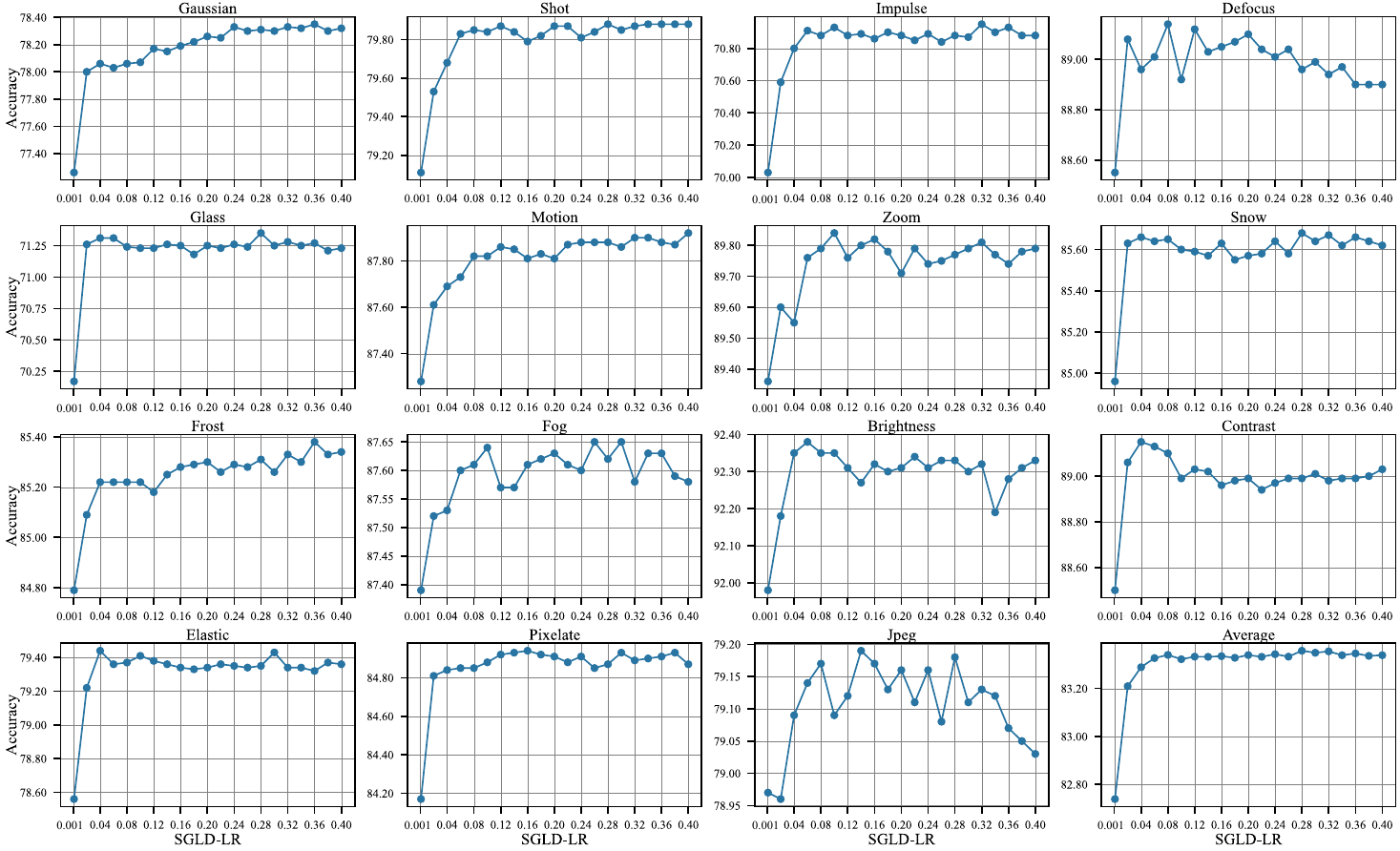}
    \caption{Hyper-parameter stability with respect to the Stochastic Gradient Langevin Dynamics (SGLD) learning rate. The x-axis is the SGLD learning rate varying from 0.001 to 0.4, while the y-axis measures model performance in terms of accuracy.}
    \label{fig:app_exp_hyper1}
    \includegraphics[width=1\linewidth]{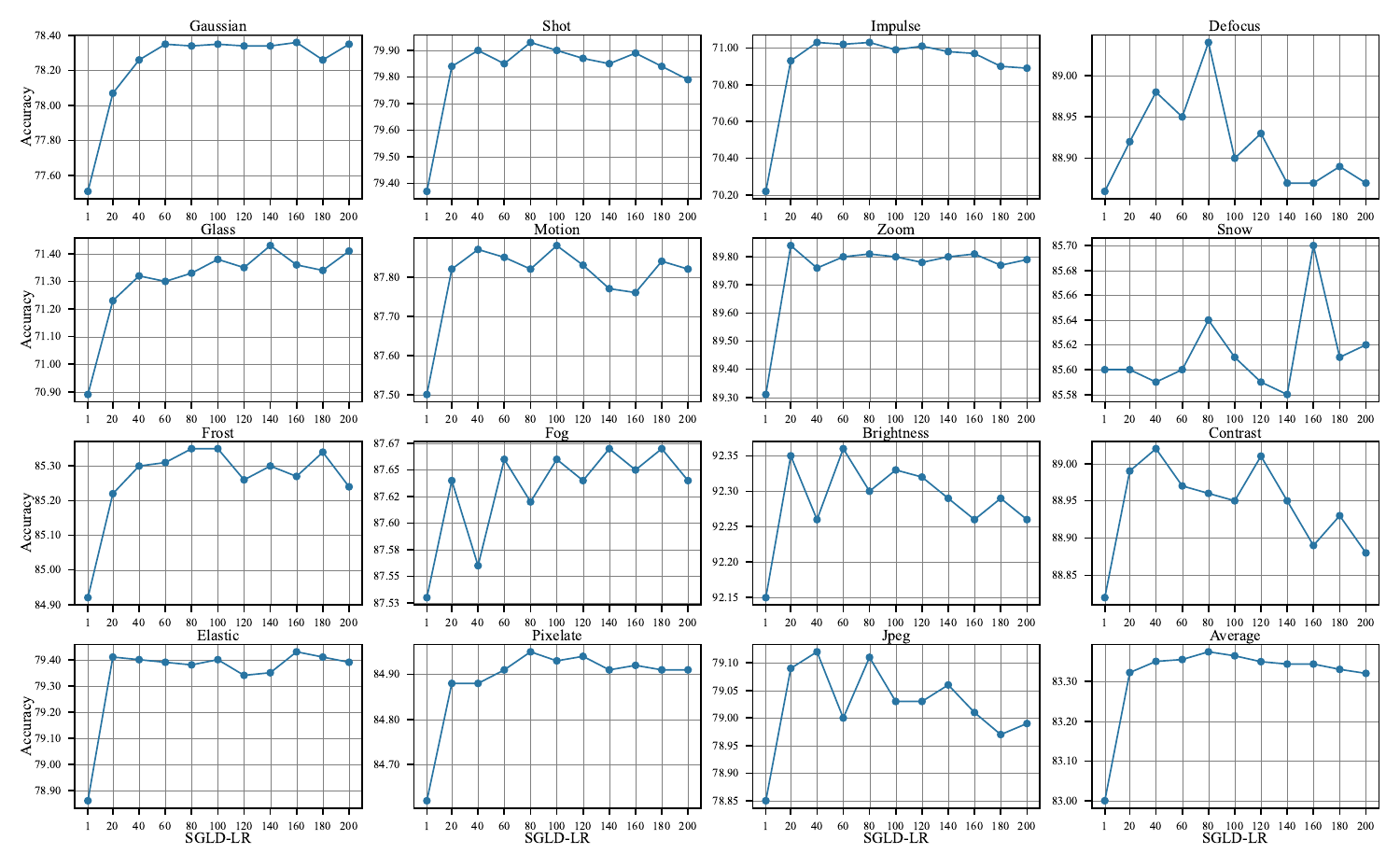}
    \caption{Hyper-parameter stability with respect to the Stochastic Gradient Langevin Dynamics (SGLD) step. The x-axis is the SGLD step varying from 1 to 200, while the y-axis measures model performance in terms of accuracy.}
    \label{fig:app_exp_hyper2}
\end{figure*}

This section provide a new experiments on hyper-parameters sensitivity of our proposed TEA.
The main hyper-parameters for TEA are the step and learning rate for Stochastic Gradient Langevin Dynamics (SGLD). 
\cref{fig:app_exp_hyper1} illustrates the variation in model accuracy as the SGLD learning rate is incrementally adjusted from 0.001 to 0.4, 
while~\cref{fig:app_exp_hyper2} demonstrates the impact on accuracy when the SGLD step is increased from 1 to 200.
The results reveal that the performance of TEA is consistently state-of-the-art under a wide range of hyper-parameters choices, across all types of corruption on CIFAR-10-C.

\section{Detailed Settings}
\label{app:setting}

\subsection{Datasets}
\label{app:datasets}

We perform experiments on four datasets across two tasks. Image corruption task include CIFAR-10(C), CIFAR-100(C), and Tiny-ImageNet(C) datasets. 
Domain generalization task include PACS datasets.

\paragraph{Dataset of Clean Distribution}
Clean distribution of CIFAR-10, CIFAR-100~\cite{krizhevsky2009learning} and Tiny-ImageNet\cite{le2015tiny} are datasets of clean distribution. CIFAR-10 and CIFAR-100 datasets consist of 60,000 color images, each of size 3x32x32 pixels. CIFAR-10 is categorized into 10 distinct classes with 6000 images per class. CIFAR-100 is more challenging, as these images are distributed across 100 classes, with 600 images per class. 
Tiny-ImageNet datasets consist of 110,000 color images, each of size 3x64x64 pixels, which are categorized into 200 distinct classes with 550 images per class.
Both CIFAR-10 and CIFAR-100 are subdivided into a training set of 50,000 images and a test set of 10,000 images. Tiny-ImageNet
is subdivided into a training set of 100,000 images and a test set of 10,000 images.

\begin{table}[!h]
\centering
\caption{Summary of Clean \& Corruption Datasets}
\begin{adjustbox}{width=\linewidth}
\setlength{\tabcolsep}{2mm}
\begin{tabular}{lccccc}
\toprule
\textbf{Dataset} & \textbf{\#Train} & \textbf{\#Test} & \textbf{\#Corr.} & \textbf{\#Severity} & \textbf{\#Class.} \\
\midrule
CIFAR-10 & 50,000 & 10,000 & 1 & 1 & 10  \\
CIFAR-100 & 50,000 & 10,000 & 1 & 1 & 100  \\
Tiny-ImageNet & 100,000 & 10,000 & 1 &1 & 200\\
CIFAR-10-C & - & 950,000 & 15 &5 & 10 \\
CIFAR-100-C & - & 950,000 & 15 &5 & 100  \\
Tiny-ImageNet-C &-& 750,000 & 15&5 & 200\\
\bottomrule
\end{tabular}
\end{adjustbox}
\label{Tab:Dataset_Corr}

\caption{Summary of PACS Datasets}
\begin{adjustbox}{width=\linewidth}
\setlength{\tabcolsep}{5mm}
\begin{tabular}{lccccc}
\toprule
\textbf{Domain} & \textbf{\#Sample} & \textbf{\#Class} & \textbf{Size} \\
\midrule
\textbf{P}hoto & 1,670 & 7 &3x227x227 \\
\textbf{A}rt & 2,048 & 7 & 3x227x227\\
\textbf{C}artoon & 2,344 & 7 & 3x227x227\\
\textbf{S}ketch & 3,929 & 7 & 3x227x227\\
\bottomrule
\end{tabular}
\end{adjustbox}
\label{Tab:Dataset_PACS}

\end{table}

\paragraph{Dataset of Corrupted Distributions}
CIFAR-10-C, CIFAR-100-C and Tiny-ImageNet-C\cite{hendrycks2018benchmarking} are variants of the original CIFAR-10, CIFAR-100 and Tiny-ImageNet datasets that have been artificially corrupted into 19 types of corruptions at five levels of severity, resulting in 95 corrupted versions of the original test set images. 
The corruptions include 15 main corruptions: Gaussian noise, shot noise, impulse noise, defocus blur, glass blur, motion blur, zoom blur, snow, frost, fog, brightness, contrast, elastic, pixelation, and JPEG.
All these corruptions are simulations of shifted distributions that models might encounter in real-world situations.

\paragraph{Datsset of PACS}
PACS\cite{li2017deeper} is an image dataset popular used in transfer learning, which consist of four domains, namely Photo (1,670 images), Art Painting (2,048 images), Cartoon (2,344 images) and Sketch (3,929 images). Each domain contains seven categories.

\subsection{Evaluation Metrics}
\label{app:metrics}
For evaluation on corruption datasets, we employ Average Accuracy and Mean Corruption Error (mCE)~\cite{hendrycks2018benchmarking} as evaluation metrics. 
For clean and PACS datasets, we employ Accuracy as evaluation metric.
These metrics provide a comprehensive evaluation of a model's generalization in handling diverse distributions, thereby offering a multi-faceted perspective on model performance.
\vspace{-3mm}
\paragraph{Average Accuracy} Average Acc is the accuracy averaged over all severity levels and corruptions. Consider there are a total of $C$ corruptions, each with $S$ severities. For a model $f$, let $\mathcal{E}_{s,c}(f)$ denote the top-1 error rate on the corruption $c$ with severity level $s$ averaged over the whole test set,
\begin{equation}
    \mathrm{AverAcc}_f=1-\frac{1}{C \cdot S} \sum_{c=1}^{C} \sum_{s=1}^S \mathcal{E}_{s, c}(f).
\end{equation}
\vspace{-3mm}
\paragraph{Mean Corruption Error} mCE is a metric used to measure the performance improvement of model $f$ compared to a baseline model $f_0$. We use the model without adaptation as the baseline model,

\begin{equation}
\mathrm{mCE}_f=\frac{1}{C} \sum_{c=1}^C \frac{\sum_{s=1}^S \mathcal{E}_{c, s}(f)}{\sum_{s=1}^S \mathcal{E}_{c, s}\left(f_0\right)} 
\end{equation}

\subsection{Hyper-parameters}
\label{app:hyper}
This section outlines the hyper-parameters chosen for our experiments. 
These settings enable the reproducibility of the results presented in our study.
For common hyperparameters, we align with those used in Tent~\cite{wang2021tent}. 
For TEA-specific hyper-parameters, we adjust them following the parameter choices from JEM~\cite{Grathwohl2020Your}.

\begin{table}[t]
\renewcommand{\arraystretch}{1}
\caption{Summary of Hyper-parameters}
\centering
\begin{adjustbox}{width=\linewidth}
\setlength{\tabcolsep}{2mm}
\begin{tabular}{cccccccc}
\toprule
\multirow{2}{*}{\textbf{Data}} & \multicolumn{4}{c}{\textbf{Common}} & \multicolumn{3}{c}{\textbf{TEA-SGLD}} \\
\cmidrule(lr){2-5} \cmidrule(lr){6-8}
& \textbf{Step} & \textbf{LR} & \textbf{BS} & \textbf{Optim} & \textbf{Step} & \textbf{LR} & \textbf{Std} \\
\midrule
CIFAR-10-C      & 1 & 0.001  & 200  & Adam & 20 & 0.1 & 0.01  \\
CIFAR-100-C     & 1 & 0.001  & 200  & Adam & 20 & 0.1 & 0.01  \\
Tiny-ImageNet-C & 1 & 0.0001 & 1000 & Adam & 20 & 0.1 & 0.01  \\
PACS-P          & 10 & 0.001 & full & Adam & 20 & 0.1 & 0.01  \\
PACS-A          & 10 & 0.001 & full & Adam & 20 & 0.1 & 0.01  \\
PACS-C          & 10 & 0.002 & full & Adam & 20 & 0.1 & 0.01  \\
PACS-S          & 20 & 0.002 & full & Adam & 20 & 0.1 & 0.01  \\
\bottomrule
\end{tabular}
\end{adjustbox}
\end{table}

\subsection{Computing resources}
\label{app:resources}
All our experiments are performed on RedHat server (4.8.5-39) with Intel(R) Xeon(R) Gold 5218 CPU $@$ 2.30GHz4, $4 \times$ NVIDIA Tesla V100 SXM2 (32GB) and $3 \times$ NVIDIA Tesla A800 SXM4 (80GB).

\section{Limitation and Future Works}
\label{app:limitation}

Our study has identified key aspects for improvement and future research, which are outlined below:
(1) The use of Stochastic Gradient Langevin Dynamics sampling is both time-consuming and unstable. However, ongoing research in energy-based models is addressing these issues through various methods, such as gradient clipping~\cite{yang2021jem++}, diffusion process~\cite{luo2023training}, additional gradient term~\cite{pmlr-v139-du21b} and ordinary differential equation based sampling~\cite{nie2021controllable}. 
One of our future directions is to enhance TEA by incorporating these advanced sampling techniques.
(2) Overemphasizing the model's sensitivity to the data distribution may significantly impact its discriminative ability. 
This trade-off between transferability and discriminability is a common theme in TTA research~\cite{kundu2022balancing, Gao_2023_CVPR}. 
Another direction for our future work is to explore how to enhance the model's perception of data  distribution while maintaining or even improving its discriminative power.
We acknowledgethat the limitations identified may present challenges.
Nevertheless, we remain confident that our study represents a pioneering effort to integrate energy-based training into test-time adaptation.  
We believe that any future advancements in the training of energy-based models will likely enhance and refine the outcomes we have demonstrated in our research.

\begin{sidewaystable*}[t] %
\renewcommand{\arraystretch}{0.86} 
\caption{
Comprehensive comparison of TEA and various baseline models on CIFAR-10-C, CIFAR-100-C and Tiny-ImageNet-C. 
All evaluated models employ the architecture of WRN-28-10 with BatchNrom.
Model weights are sourced from RobustBench for CIFAR-10-C to ensure a fair comparison.
Evaluations are based on Accuracy (\%) for each individual corruption, as well as Average Accuracy (Acc \%), Mean Corruption Error (mCE \%) for overall performance. 
The reported performance of our TEA reflects the average across five runs with varying seeds, with a maximum standard deviation under 0.1\%.
The most notable results are indicated with \textbf{boldface} for the top performance.}
\centering
\begin{small}
\begin{adjustbox}{width=\linewidth}
\setlength{\tabcolsep}{1.7mm}{
\begin{tabular}{clccccccccccccccc|cc}
\toprule
& \multirow{2}*{Method} & \multicolumn{3}{c}{Noise} & \multicolumn{4}{c}{Blur} & \multicolumn{3}{c}{Weather} & \multicolumn{5}{c}{Digital} & \multicolumn{2}{c}{Avg} \\
\cmidrule(lr){3-5} \cmidrule(lr){6-9} \cmidrule(lr){10-12} \cmidrule(lr){13-17} \cmidrule(lr){18-19}
& & Gaussian & Shot & Impulse & Defocus & Glass & Motion & Zoom & Snow & Frost & Fog & Bright & Contrast & Elastic & Pixel & JPEG & Acc($\uparrow$) & mCE($\downarrow$)  \\
\midrule
\multirow{10}{*}{\rotatebox{90}{CIFAR-10(-C)}} 
& \cellcolor{pink!20}Source & 27.68 & 34.25 & 27.07 & 53.01 & 45.67 & 65.24 & 57.99 & 74.87 & 58.68 & 73.98 & 90.70 & 53.37 & 73.39 & 41.56 & 69.71 & 56.47 & 100.00 \\
& BN & 71.93 & 73.88 & 63.74 & 87.19 & 64.72 & 85.83 & 87.89 & 82.73 & 82.61 & 84.75 & 91.61 & 87.35 & 76.25 & 80.33 & 72.70 & 79.56 & 52.65 \\
& DUA* & 72.60 & 75.40 & 64.70 & 86.90 & 65.10 & 85.40 & 88.40 & 83.20 & 82.50 & 86.90 & \textbf{92.40} & 85.90 & 77.30 & 80.70 & 73.80 & 80.10 & 50.78 \\
& PL & 14.46 & 17.77 & 14.48 & 44.27 & 31.29 & 66.70 & 60.49 & 76.46 & 55.01 & 79.35 & 91.05 & 43.49 & 75.36 & 31.61 & 69.63 & 51.42 & 106.98\\
& SHOT & 62.91 & 66.15 & 46.41 & 82.85 & 61.64 & 82.39 & 83.16 & 79.03 & 78.24 & 82.35 & 89.68 & 82.20 & 76.08 & 75.34 & 73.19 & 74.77 & 63.19 \\
& TENT & 75.20 & 76.52 & 67.03 & 88.00 & 68.22 & 86.27 & 89.24 & 84.06 & 83.81 & 86.30 & 92.16 & 87.91 & 78.01 & 82.77 & 75.77 & 81.41 & 48.13\\
& ETA & 72.21 & 73.88 & 63.72 & 87.21 & 64.70 & 85.84 & 87.89 & 82.73 & 82.58 & 84.77 & 91.60 & 87.32 & 76.22 & 80.37 & 72.69 & 79.58 & 52.64\\
& EATA & 72.25 & 73.88 & 63.74 & 87.19 & 64.72 & 85.83 & 87.89 & 82.73 & 82.61 & 84.75 & 91.61 & 87.35 & 76.25 & 80.33 & 72.70 & 79.59 & 52.62\\
& SAR & 71.95 & 74.14 & 64.11 & 87.39 & 65.20 & 86.00 & 88.06 & 83.08 & 82.66 & 85.07 & 91.90 & 87.20 & 76.69 & 80.41 & 72.79 & 79.77 & 51.94\\
\cmidrule(lr){2-19}
& \cellcolor{gray!20} TEA & \cellcolor{gray!20} \textbf{78.33} & \cellcolor{gray!20} \textbf{79.87} & \cellcolor{gray!20} \textbf{70.94} & \cellcolor{gray!20} \textbf{88.89} & \cellcolor{gray!20} \textbf{71.31} & \cellcolor{gray!20} \textbf{87.87} & \cellcolor{gray!20} \textbf{89.77} & \cellcolor{gray!20} \textbf{85.56} & \cellcolor{gray!20} \textbf{85.29} & \cellcolor{gray!20} \textbf{87.61} & \cellcolor{gray!20} 92.37 & \cellcolor{gray!20} \textbf{88.98} & \cellcolor{gray!20} \textbf{79.32} & \cellcolor{gray!20} \textbf{84.90} & \cellcolor{gray!20} \textbf{78.99} & \cellcolor{gray!20} \textbf{83.33} & \cellcolor{gray!20} \textbf{43.69} \\
\midrule
\multirow{9}{*}{\rotatebox{90}{CIFAR-100(-C)}} 
& \cellcolor{pink!20}Source & 9.87 & 11.58 & 4.15 & 35.57 & 20.56 & 46.92 & 44.20 & 51.13 & 39.07 & 45.07 & 71.42 & 27.43 & 51.01 & 30.19 & 42.82 & 35.39 & 100.00\\
& BN & 46.53 & 48.62 & 37.15 & 70.94 & 47.36 & 69.04 & 71.25 & 63.00 & 62.96 & 66.08 & 75.89 & 71.31 & 58.79 & 64.56 & 47.46 & 60.06 & 63.54\\
& PL & 29.54 & 34.09 & 14.99 & 64.06 & 40.81 & 65.36 & 67.74 & 60.42 & 59.47 & 62.92 & 75.98 & 57.29 & 58.72 & 59.41 & 50.34 & 53.40 & 72.12\\
& SHOT & 37.50 & 39.68 & 21.27 & 67.90 & 45.52 & 67.42 & 69.86 & 61.42 & 60.75 & 65.01 & 75.12 & 63.96 & 58.96 & 62.54 & 51.09 & 56.53 & 68.01\\
& TENT & 53.44 & 54.12 & 45.53 & 71.69 & 51.17 & 71.54 & 71.63 & 64.88 & 65.30 & 68.41 & 75.14 & 73.59 & 59.25 & 66.81 & 53.93 & 63.09 & 59.42\\
& ETA & 48.64 & 50.95 & 38.05 & 69.66 & 47.52 & 67.76 & 70.24 & 62.51 & 61.73 & 66.03 & 73.40 & 71.15 & 56.93 & 64.40 & 48.34 & 59.82 & 64.52\\
& EATA & 48.76 & 51.60 & 39.47 & 69.29 & 47.49 & 68.13 & 70.68 & 62.94 & 62.53 & 65.14 & 74.48 & 71.61 & 57.53 & 64.24 & 49.67 & 60.24 & 63.75\\
& SAR & 51.34 & 54.08 & 44.62 & 72.24 & 50.10 & 71.06 & 72.43 & 64.96 & 65.35 & 68.40 & 76.23 & 73.95 & 60.04 & 67.26 & 52.29 & 62.95 & 59.37\\
\cmidrule(lr){2-19}
& \cellcolor{gray!20} TEA & \cellcolor{gray!20} \textbf{54.29} & \cellcolor{gray!20} \textbf{56.55} & \cellcolor{gray!20} \textbf{48.59} & \cellcolor{gray!20} \textbf{72.96} & \cellcolor{gray!20} \textbf{53.78} & \cellcolor{gray!20} \textbf{72.63} & \cellcolor{gray!20} \textbf{74.20} & \cellcolor{gray!20} \textbf{67.78} & \cellcolor{gray!20} \textbf{67.14} & \cellcolor{gray!20} \textbf{69.98} & \cellcolor{gray!20} \textbf{76.74} & \cellcolor{gray!20} \textbf{75.71} & \cellcolor{gray!20} \textbf{62.18} & \cellcolor{gray!20} \textbf{68.65} & \cellcolor{gray!20} \textbf{55.32} & \cellcolor{gray!20} \textbf{65.10} & \cellcolor{gray!20} \textbf{56.18} \\
\midrule
\multirow{9}{*}{\rotatebox{90}{Tiny-ImageNet(-C)}} 
& \cellcolor{pink!20}Source & 12.37 & 16.07 & 5.55 & 6.63 & 5.67 & 20.71 & 19.38 & 26.79 & 31.88 & 14.18 & 29.75 & 1.95 & 32.03 & 49.00 & 46.19 & 21.21 & 100.00 \\
& BN & 24.00 & 25.82 & 18.22 & 28.43 & 18.46 & 35.20 & 33.41 & 27.59 & 30.26 & 25.58 & 32.58 & 6.04 & 34.12 & 40.81 & 35.72 & 27.74 & 93.42 \\
& PL & 20.75 & 25.99 & 11.41 & 17.28 & 14.42 & 36.07 & 33.66 & 30.19 & 35.01 & 22.44 & 34.87 & 2.03 & \textbf{41.31} & 51.76 & 46.75 & 28.26 & 91.22 \\
& SHOT & 22.25 & 27.10 & 14.54 & 18.74 & 15.17 & 36.78 & 34.63 & 30.95 & \textbf{35.68} & 23.52 & 36.01 & 1.91 & 41.01 & \textbf{51.99} & \textbf{46.85} & 29.14 & 90.16\\
& TENT & 23.80 & 25.29 & 18.68 & 27.05 & 17.54 & 33.21 & 31.36 & 26.21 & 28.56 & 24.76 & 30.08 & 5.68 & 31.37 & 37.77 & 33.33 & 26.31 & 95.52\\
& ETA & 24.44 & 25.76 & 18.00 & 28.05 & 18.25 & 34.59 & 32.85 & 27.11 & 29.87 & 25.42 & 31.86 & 5.93 & 32.84 & 39.55 & 34.82 & 27.28 & 94.12\\
& EATA & 24.48 & 25.55 & 17.95 & 27.68 & 18.41 & 34.83 & 32.84 & 27.03 & 29.67 & 25.39 & 31.94 & 5.87 & 32.71 & 39.87 & 35.09 & 27.28 & 94.09\\
& SAR & 25.10 & 26.33 & 18.87 & 28.80 & 18.44 & 35.59 & 33.87 & 28.31 & 30.62 & 26.32 & 33.31 & 6.30 & 34.24 & 41.16 & 35.97 & 28.21 & 92.82\\
\cmidrule(lr){2-19}
& \cellcolor{gray!20} TEA & \cellcolor{gray!20} \textbf{25.50} & \cellcolor{gray!20} \textbf{29.35} & \cellcolor{gray!20} \textbf{20.58} & \cellcolor{gray!20} \textbf{32.76} & \cellcolor{gray!20} \textbf{21.14} & \cellcolor{gray!20} \textbf{41.72} & \cellcolor{gray!20} \textbf{39.06} & \cellcolor{gray!20} \textbf{31.43} & \cellcolor{gray!20} 34.19 & \cellcolor{gray!20} \textbf{29.02} & \cellcolor{gray!20} \textbf{36.83} & \cellcolor{gray!20} \textbf{6.89} & \cellcolor{gray!20} 38.94 & \cellcolor{gray!20} 46.38 & \cellcolor{gray!20} 41.31 & \cellcolor{gray!20} \textbf{31.67} & \cellcolor{gray!20} \textbf{87.99}\\
\bottomrule
\end{tabular}}
\end{adjustbox}
\end{small}
\label{tab:app_mainada_full}
\end{sidewaystable*}

\end{document}